\documentclass[10pt,twocolumn,letterpaper]{article}

\usepackage[pagenumbers]{cvpr} 
\usepackage{adjustbox} 

\usepackage[ruled]{algorithm2e} 

\SetAlFnt{\small}
\SetAlCapFnt{\small}
\SetAlCapNameFnt{\small}
\SetAlCapHSkip{0pt}

\definecolor{cvprblue}{rgb}{0.21,0.49,0.74}

\usepackage{graphicx}
\usepackage{amsmath}
\usepackage{amssymb}
\usepackage{booktabs}

\usepackage{color}
\usepackage{makecell}

\definecolor{tabhigh}{rgb}{0.9, 0.25, 0.25}
\definecolor{tabmid}{rgb}{0.9, 0.6, 0.05}
\definecolor{tablow}{rgb}{0.4, 0.8, 0.35}

\usepackage{tikz}
\usepackage{pgfplots}
\pgfplotsset{compat=1.18}
\usepackage{graphicx}
\usepackage{amsmath}
\usepackage{amssymb}
\usepackage{booktabs}
\usepackage{url}
\usepackage{epsfig}
\usepackage{lipsum}
\usepackage{amsthm}
\usepackage{comment}
\usepackage{multirow}
\usepackage{subcaption}
\usepackage{microtype}
\usepackage{xspace}
\usepackage{enumitem}
\usepackage{graphbox}
\usepackage{epigraph}
\usepackage{wrapfig}
\usepackage{textcomp}  
\usepackage{scalerel}  
\usepackage{animate}
\usepackage{colortbl}
\usepackage{tabularx,booktabs}

\usepackage{multirow}
\usepackage{dsfont}
\usepackage{enumitem}

\usepackage{algorithm}
\usepackage{algpseudocode}
\usepackage{animate}
\usepackage{booktabs} 
\usepackage{hhline}
\usepackage{adjustbox}
\usepackage{multirow}
\usepackage{booktabs}
\usepackage[normalem]{ulem}
\usepackage{ifthen}
\usepackage{epsfig}
\usepackage{graphics}
\usepackage{graphicx}
\usepackage{amsmath}
\usepackage{animate}
\usepackage{amsmath}
\usepackage{booktabs}
\usepackage{hhline}
\usepackage{adjustbox}
\usepackage{booktabs,arydshln}
\usepackage{amsfonts}
\usepackage{color}

\definecolor{cvprblue}{rgb}{0.21,0.49,0.74}
\usepackage[pagebackref,breaklinks,colorlinks,allcolors=cvprblue]{hyperref}

\def\paperID{8738} 
\def\confName{CVPR}
\def\confYear{2026}

\title{DreamLoop: Controllable Cinemagraph Generation from a Single Photograph}

\author{
  Aniruddha Mahapatra\ \ \
  Long Mai\ \ \ 
  Cusuh Ham \ \ \
  Feng Liu\\[1ex]
    Adobe Research\\[1ex]
\url{https://anime26398.github.io/dreamloop.github.io/}
}

\begin{document}

\newcommand{\reffig}[1]{Figure~\ref{fig:#1}}
\newcommand{\refsec}[1]{Section~\ref{sec:#1}}
\newcommand{\refapp}[1]{Appendix~\ref{sec:#1}}
\newcommand{\reftbl}[1]{Table~\ref{tab:#1}}
\newcommand{\refalg}[1]{Algorithm~\ref{alg:#1}}
\newcommand{\refline}[1]{Line~\ref{line:#1}}
\newcommand{\shortrefsec}[1]{\S~\ref{sec:#1}}
\newcommand{\refeq}[1]{Equation~\ref{eq:#1}}
\newcommand{\refeqshort}[1]{(\ref{eq:#1})}
\newcommand{\shortrefeq}[1]{\ref{eq:#1}}
\newcommand{\lblfig}[1]{\label{fig:#1}}
\newcommand{\lblsec}[1]{\label{sec:#1}}
\newcommand{\lbleq}[1]{\label{eq:#1}}
\newcommand{\lbltbl}[1]{\label{tab:#1}}
\newcommand{\lblalg}[1]{\label{alg:#1}}
\newcommand{\lblline}[1]{\label{line:#1}}

\definecolor{MyDarkBlue}{rgb}{0,0.08,1}
\newcommand{\camera}[1]{#1}

\newcommand{\myparagraph}[1]{\noindent\textbf{#1}}
\newcommand{\feature}{\bs{f}_{t}\xspace}

\newcommand{\website}{\href{https://anime26398.github.io/dreamloop.github.io/}{\textit{website}}\xspace}



\newcommand{\bs}[1]{{\boldsymbol{#1}}}
\newcommand{\pixel}{p\xspace} 
\newcommand{\art}{\bs{x}\xspace} 
\newcommand{\nat}{\hat{\bs{x}}\xspace} 
\newcommand{\artt}{\bs{x}_t\xspace} 
\newcommand{\natt}{\hat{\bs{x}}_t\xspace}  
\newcommand{\arttminus}{\bs{x}_{t-1}\xspace}  
\newcommand{\nattminus}{\hat{\bs{x}}_{t-1}\xspace}  
\newcommand{\atten}{\bs{A}_t\xspace}  
\newcommand{\attenavg}{\bs{\overline{A}}\xspace}  
\newcommand{\artmask}{\bs{M}\xspace}   
\newcommand{\natmask}{\hat{\bs{M}}\xspace}  
\newcommand{\artflow}{\bs{F}\xspace}  
\newcommand{\natflow}{\hat{\bs{F}}\xspace} 

\newcommand{\at}{\bs{c}\xspace}   
\newcommand{\nt}{\hat{\bs{c}}\xspace} 
\newcommand{\gf}{\bs{G}_{flow}\xspace}
\newcommand{\gv}{\bs{G}_{frame}\xspace}

\newcommand{\magvit}{\text{MagViT-v2}\xspace}
\newcommand{\lio}{\text{ProMAG}\xspace}
\newcommand{\fourx}{4$\times$\xspace}
\newcommand{\eightx}{8$\times$\xspace}
\newcommand{\sixteenx}{16$\times$\xspace}


\newcommand{\dreamloop}{DreamLoop\xspace}



\definecolor{myblue}{rgb}{0.239,0.553,0.565}

\definecolor{cmu_red}{rgb}{0.706,0.169,0.212}

\newlength{\cellw}
\newlength{\cellh}
\newlength{\rowgap}
\setlength{\cellw}{0.24\textwidth} 
\setlength{\cellh}{6.2cm}          
\setlength{\rowgap}{0.6em}          
\setlength{\tabcolsep}{0pt}         
\setlength{\fboxrule}{0.8pt}
\setlength{\fboxsep}{0pt}
\setkeys{Gin}{draft=false}          
\newlength{\innergap}\setlength{\innergap}{0.2em}   
\newlength{\outergap}\setlength{\outergap}{0.1em}   
\newlength{\leftw}\setlength{\leftw}{\dimexpr 2\cellw + \innergap\relax} 

\twocolumn[{%
  \maketitle
  \begin{center}
    \begin{minipage}{\textwidth}
      \centering
        \begin{tabular}[t]{@{} c @{\hspace{\outergap}} c @{}} 
          \begin{minipage}[t]{0.25\textwidth}
            \vspace{0pt}
            \includegraphics[width=\linewidth,
                             height=\dimexpr 2\cellh + \rowgap\relax,
                             keepaspectratio]{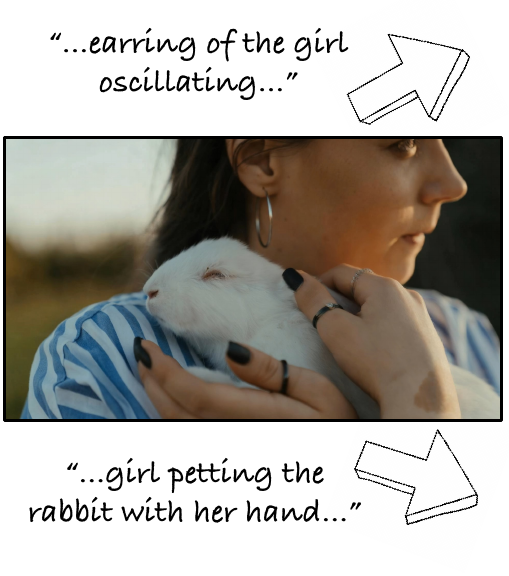}
          \end{minipage}
          &
          \begin{tabular}[t]{@{} c @{\hspace{\innergap}} c @{\hspace{\innergap}} c @{}}
            \begin{minipage}[t]{\cellw}
              \vspace{0pt}
              \includegraphics[width=\linewidth,height=\cellh,keepaspectratio]{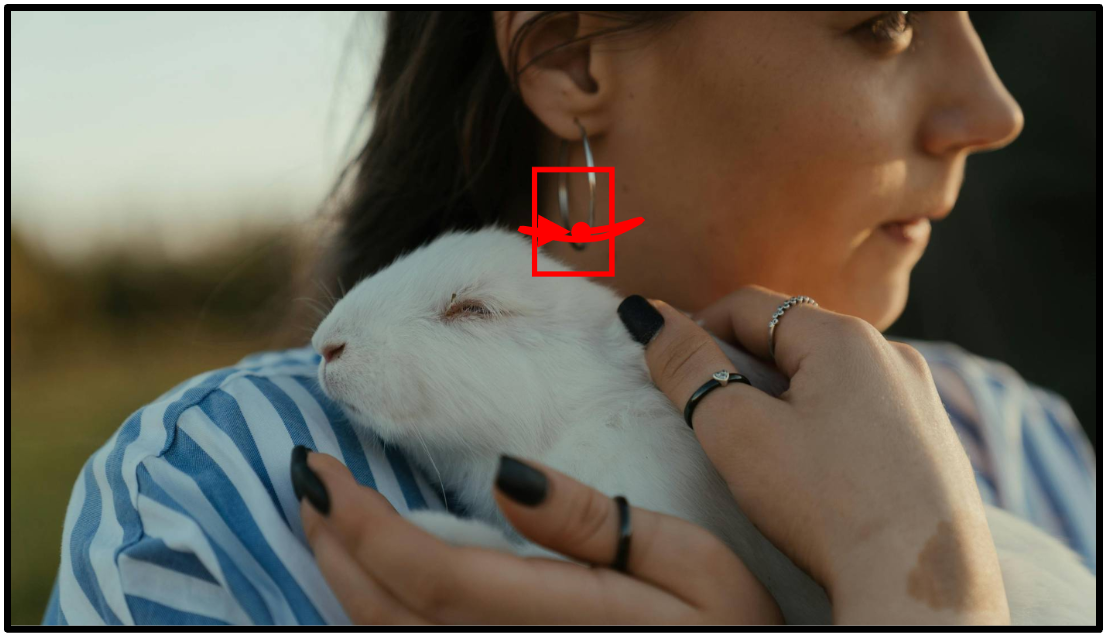}
            \end{minipage} &
            \begin{minipage}[t]{\cellw}
              \vspace{0pt}
              \fcolorbox{black}{white}{%
                \animategraphics[autoplay,loop,width=\linewidth]{24}{images/teaser/teaser-1/}{0}{71}%
              }
            \end{minipage} &
            \begin{minipage}[t]{\cellw}
              \vspace{0pt}
              \includegraphics[width=\linewidth,height=\cellh,keepaspectratio]{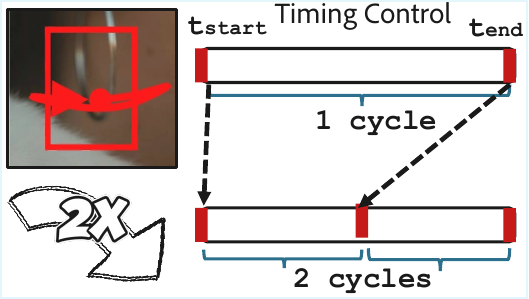}
            \end{minipage} \\[\rowgap]
            \begin{minipage}[t]{\cellw}
              \vspace{0pt}
              \includegraphics[width=\linewidth,height=\cellh,keepaspectratio]{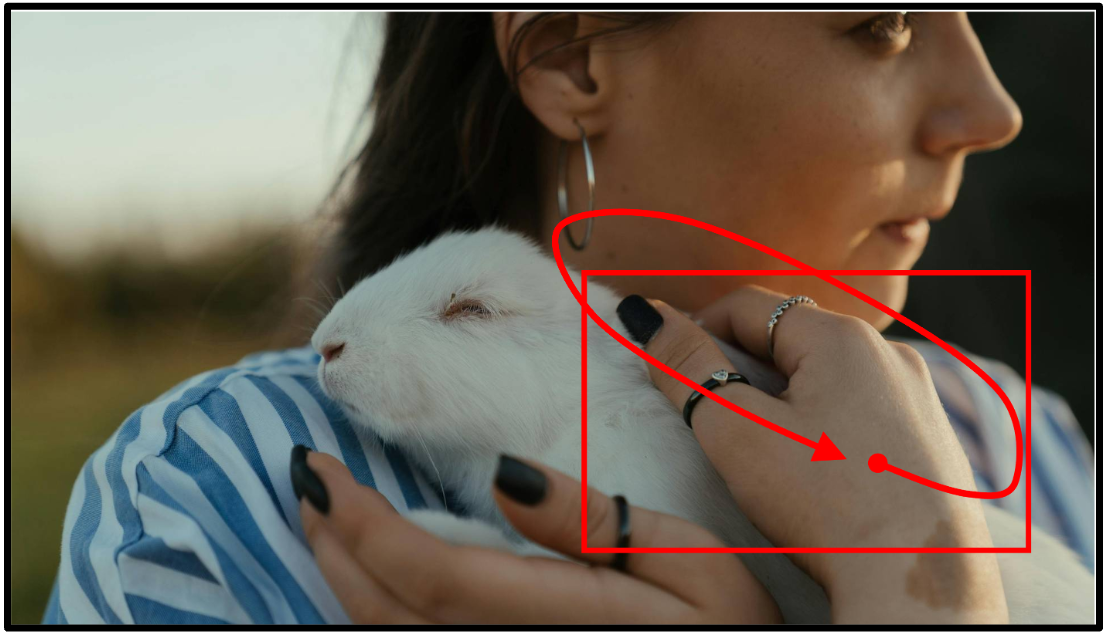}
            \end{minipage} &
            \begin{minipage}[t]{\cellw}
              \vspace{0pt}
              \fcolorbox{black}{white}{%
                \animategraphics[autoplay,loop,width=\linewidth]{24}{images/teaser/teaser-2/}{0}{119}%
              }
            \end{minipage} &
            \begin{minipage}[t]{\cellw}
              \vspace{0pt}
              \fcolorbox{black}{white}{%
                \animategraphics[autoplay,loop,width=\linewidth]{24}{images/teaser/teaser-3/}{0}{71}%
              }
            \end{minipage}
          \end{tabular}
        \end{tabular}
      \vspace{0.5em}
      \captionof{figure}{\textbf{Teaser.} Starting with the same input photograph, we showcase \textit{\dreamloop's} capability to let user generate diverse cinemagraphs with precise motion control. (top-middle) the user intends to animate the earring of the girl, oscillating. (bottom-middle) the user simulates the action of `girl petting the rabbit'. \textit{\dreamloop's} generates high-quality cinemagraph following user input in both cases. User can also play with timing control, (left) user doubles the oscillating frequency of the girls earring and \textit{\dreamloop} accurately captures it. \textit{We encourage readers to view the videos in the figure with Acrobat Reader.}}
      \label{fig:teaser}
    \end{minipage}
  \end{center}
}]

\begin{abstract}
Cinemagraphs, which combine static photographs with selective, looping motion, offer unique artistic appeal. Generating them from a single photograph in a controllable manner is particularly challenging. 
Existing image-animation techniques are restricted to simple, low-frequency motions and operate only in narrow domains with repetitive textures like water and smoke. In contrast, large-scale video diffusion models are not tailored for cinemagraph constraints and lack the specialized data required to generate seamless, controlled loops.
We present \textbf{\dreamloop}, a controllable video synthesis framework dedicated to generating cinemagraphs from a single photo without requiring any cinemagraph training data. Our key idea is to adapt a general video diffusion model by training it on two objectives: temporal bridging and motion conditioning. This strategy enables flexible cinemagraph generation. During inference, by using the input image as both the first- and last- frame condition, we enforce a seamless loop. By conditioning on static tracks, we maintain a static background. Finally, by providing a user-specified motion path for a target object, our method provides intuitive control over the animation's trajectory and timing. To our knowledge, DreamLoop is the first method to enable cinemagraph generation for general scenes with flexible and intuitive controls. We demonstrate that our method produces high-quality, complex cinemagraphs that align with user intent, outperforming existing approaches.

\end{abstract}    
\section{Introduction}
\label{sec:intro}

Cinemagraph is an art form that combines the appeal of static images and dynamic videos. It depicts a scene where most elements are static, while a few selected regions move in a seamless, looping manner. While this medium is popular in creative communities, its creation remains a significant challenge. The conventional process demands a carefully planned video shoot, tripod stabilization, and tedious post-production for motion segmentation and frame blending.

\noindent Generating a cinemagraph from a single photograph offers a promising way to make this art form more accessible. However, this task presents a unique set of challenges distinct from standard image-to-video (I2V) generation. First, the animation must be selective; unlike typical I2V models~\cite{yang2024cogvideox, wan2025, kong2024hunyuanvideo} that animate all plausible elements, a cinemagraph requires keeping most of the scene static to preserve its photographic nature. Second, the generated motion must be seamlessly looping to be played infinitely. Finally, the process must be highly controllable, allowing users to define which regions move, how they move (e.g., path, direction), and the timing of that motion (e.g., speed, looping frequency).

\noindent Existing single-image–to-cinemagraph methods~\cite{holynski2021animating, endo2019animating, mahapatra2023text, simo2016learning, halperin2021endless} are limited to narrow domains with repetitive textures (\textit{`fluid elements'}~\cite{mahapatra2022controllable}) and can animate only simple motions such as water, smoke, etc. 
A major limitation of these approaches is that most of them~\cite{holynski2021animating, mahapatra2022controllable, mahapatra2023text, endo2019animating, halperin2021endless} rely predominantly on optical flow prediction, resulting in \textit{in-place} motion of repeating textures; thus, they are fundamentally incapable of animating general scene elements, such as humans, animals, or everyday objects, as shown in~\reffig{teaser}.
On the other hand, large-scale video diffusion models, while powerful, are not designed for these specific constraints. They struggle to maintain a static background, do not inherently produce loops, and lack the fine-grained motion control necessary for artistic cinemagraph creation. Furthermore, training a dedicated model is hampered by the lack of large-scale, high-quality cinemagraph datasets.

\noindent In this paper, we present a novel framework for generating high-quality, controllable cinemagraph from a single image, without requiring cinemagraph data for training. Our central idea is to adapt a standard video diffusion model (VDM) trained on general video data to this specific task. We achieve this by augmenting the VDM with two key conditioning signals during training: (1) a video bridging objective, where the model learns to generate intermediate frames given the first and last frames, and (2) a motion anchoring objective, where the model learns to generate video conditioned on explicit motion signals, including bounding box sequence and sparse point tracks.

\noindent This training strategy enables flexible and fully controllable cinemagraph generation. At inference time, to achieve a seamless loop, we use the input photograph as both the first and last frames, forcing the model to generate a perfectly looping sequence. To ensure a static background, we provide static motion tracks for background regions. Finally, user control is achieved by allowing the user to specify a motion path for a selected object, which serves as the motion-conditioning signal. This allows for intuitive control over the motion trajectory and timing from a single input image. To the best of our knowledge, this is the first method to enable cinemagraph generation from a single image of a general scene with such flexible and intuitive control. We demonstrate that our method can produce high-quality, complex cinemagraphs that align with user intent, outperforming existing approaches. Despite being trained on general videos, our method even surpasses specialized cinemagraph methods on fluid-element domains such as water and smoke.

\noindent In summary, this paper makes the following contributions: 
\begin{itemize}
    \item We introduce the first method capable of generating cinemagraphs from general-purpose photographs—not limited to fluid elements—by finetuning a pretrained image-to-video diffusion model with (1) first–last frame conditioning for seamless looping and (2) motion conditioning via bounding-box sequences and sparse point tracks for precise motion control.
    \item We propose a user-friendly and intuitive control framework (and accompanying UI) that enables users to specify object trajectories, motion paths, and background constraints, enabling precise and flexible cinemagraph creation from a single input image.
\
\end{itemize}

\section{Related Work} 
\label{sec:related_work}
\myparagraph{Video Looping and Cinemagraph.}  
Classic cinemagraph creation methods typically operate on real videos containing periodic motion. The seminal Video Textures approach~\cite{schodl2000video} introduced a graph-based formulation to identify visually consistent frame transitions, enabling seamless looping. Subsequent works~\cite{kwatra2003graphcut, tompkin2011towards, bai2012selectively, joshi2012cliplets} extend this idea by decomposing a scene into dynamic and static regions, often requiring user-provided masks or careful video stabilization to isolate looping elements. Other methods~\cite{liao2013automated, liao2015fast} allow different regions to loop at independent paces, while additional efforts explore handling challenging settings such as moving cameras~\cite{sevilla2015smooth}, portraits~\cite{bai2013automatic}, and panoramic imagery~\cite{agarwala2005panoramic, he2017gigapixel}.
A key limitation of these video-based methods is that they require an input video captured with a very stable camera setup, which typically demands substantial effort, user guidance, and equipment such as tripods or controlled environments. In contrast, our method enables cinemagraph creation directly from a single photograph, greatly reducing the required user effort while still generating highly controllable animation.

\begin{figure*}[t!]
    \centering
    \includegraphics[width=\linewidth]{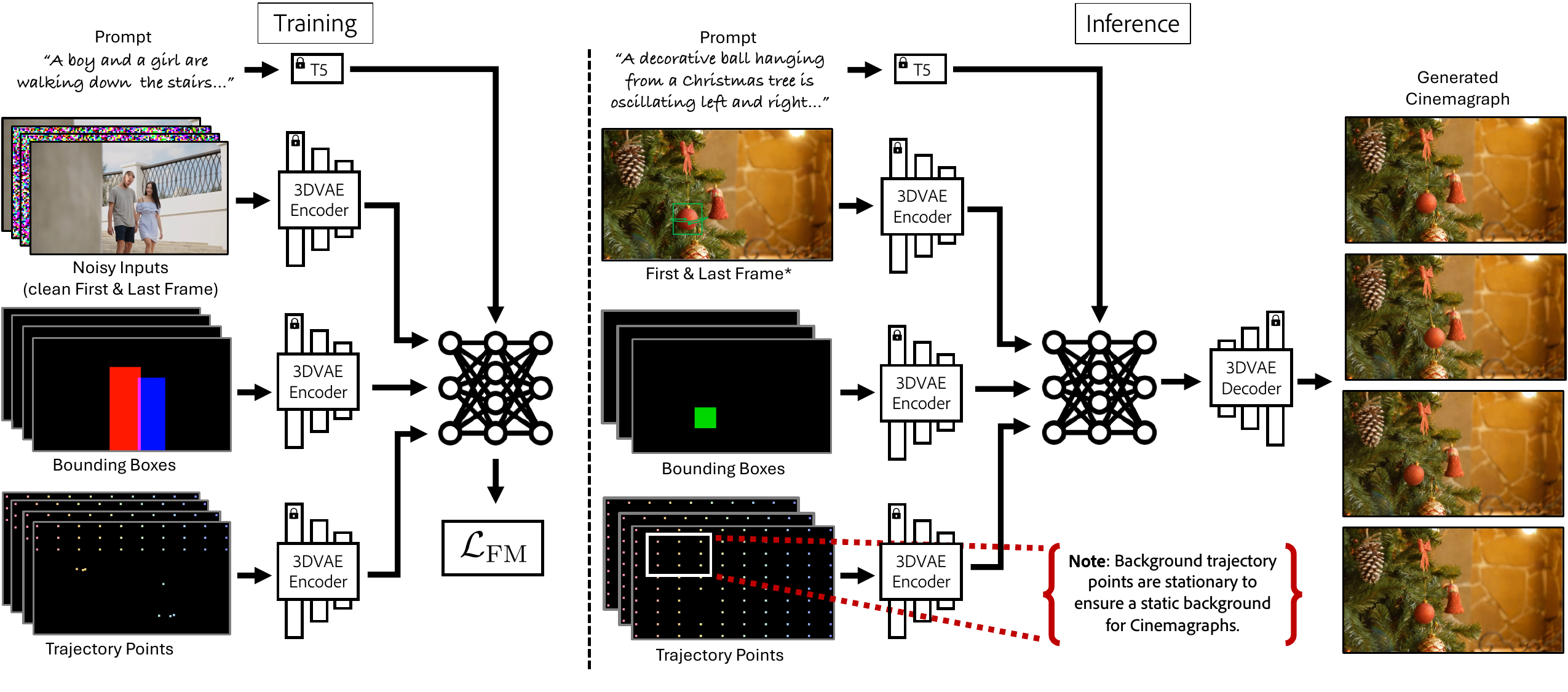}
    \caption{\camera{{\bf Methodology.} Figure shows details of our method \textit{\dreamloop}. (left) shows the training procedure with bounding box and sparse point track control. (right) shows the inference setting for generating controllable cinemagraphs with our method.}}
    \lblfig{method}
\end{figure*}

\begin{figure*}[h]
\centering
\setlength{\fboxrule}{0.8pt}
\setlength{\fboxsep}{0pt}

\begin{tabular}{@{\hspace{0em}} c @{\hspace{0.2em}} c @{\hspace{1.4em}} c @{\hspace{0.4em}} c @{\hspace{0em}}}
\begin{minipage}[t]{0.23\textwidth}
  \includegraphics[width=\linewidth]{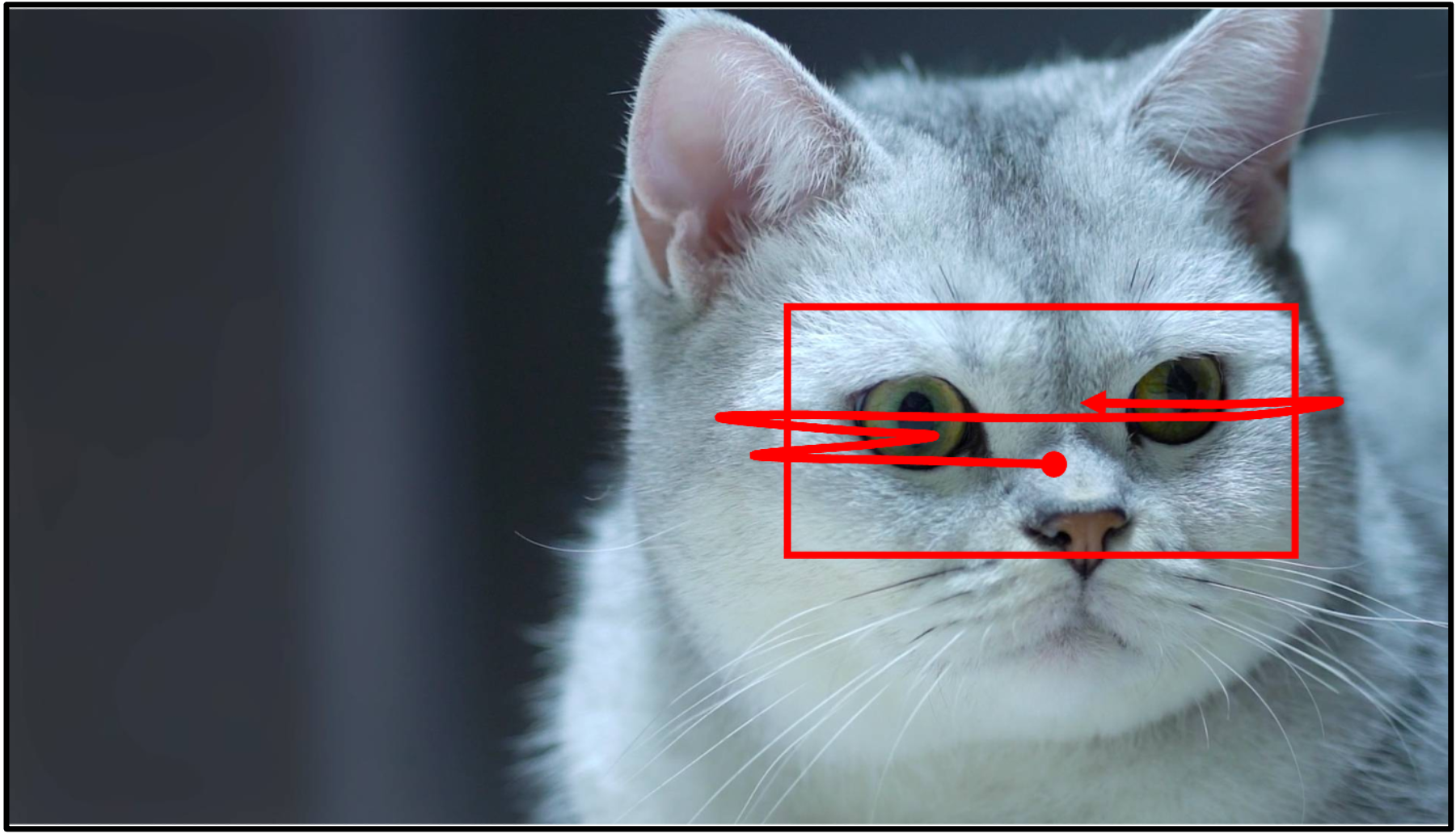}
\end{minipage} &
\begin{minipage}[t]{0.23\textwidth}
  \fcolorbox{black}{white}{\animategraphics[autoplay,loop,width=\linewidth]{24}{images/our_results/1/}{0}{71}}
\end{minipage} &
\begin{minipage}[t]{0.23\textwidth}
  \includegraphics[width=\linewidth]{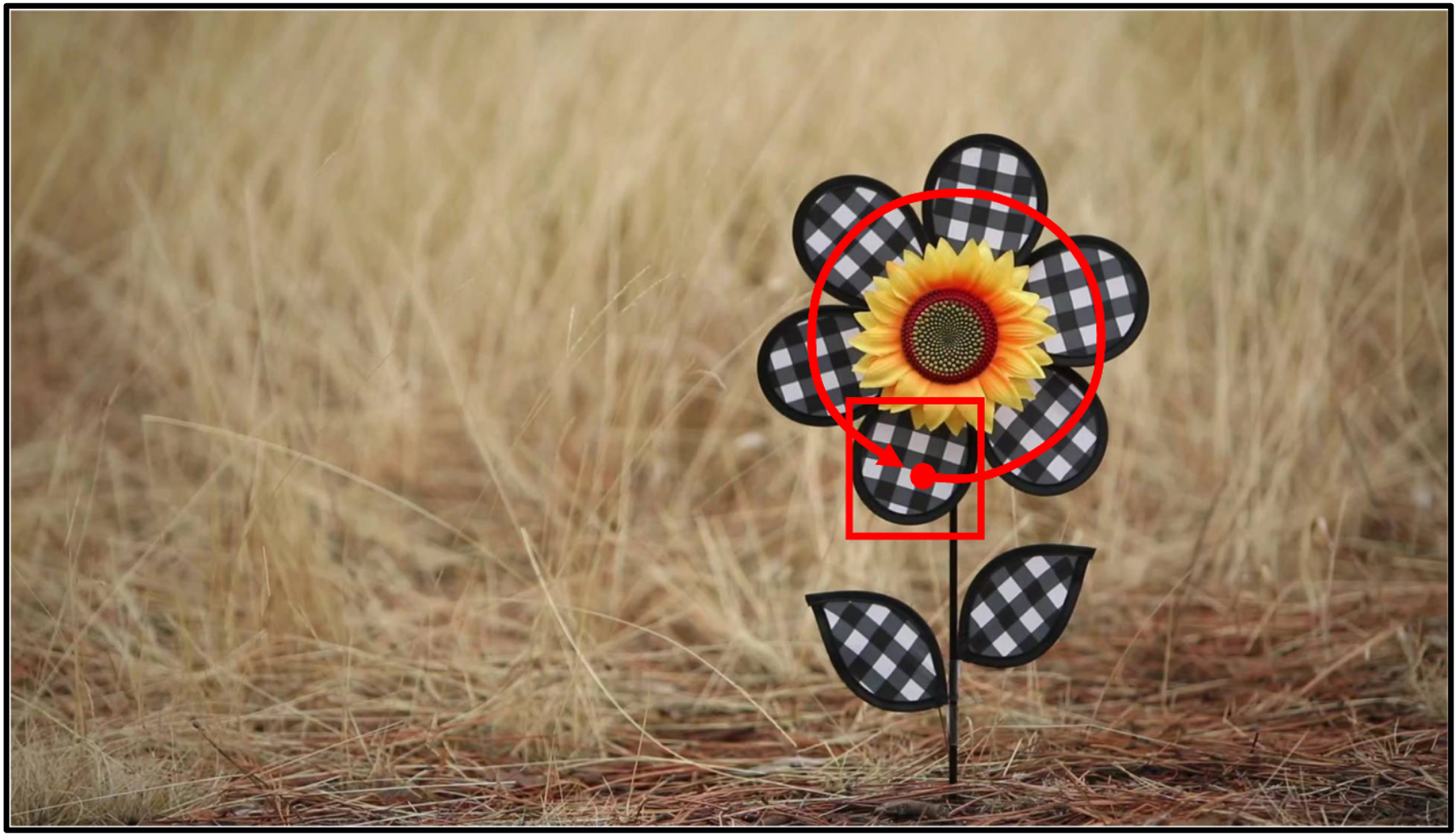}
\end{minipage} &
\begin{minipage}[t]{0.23\textwidth}
  \fcolorbox{black}{white}{\animategraphics[autoplay,loop,width=\linewidth]{24}{images/our_results/2/}{0}{71}}
\end{minipage} \\[-0.3em]
\multicolumn{2}{c}{\small (a) \texttt{`A close-up of an adorable white cat...'}} &
\multicolumn{2}{c}{\small (b) \texttt{`A plastic toy sunflower in a field...'}} \\[0.4em]

\begin{minipage}[t]{0.23\textwidth}
  \includegraphics[width=\linewidth]{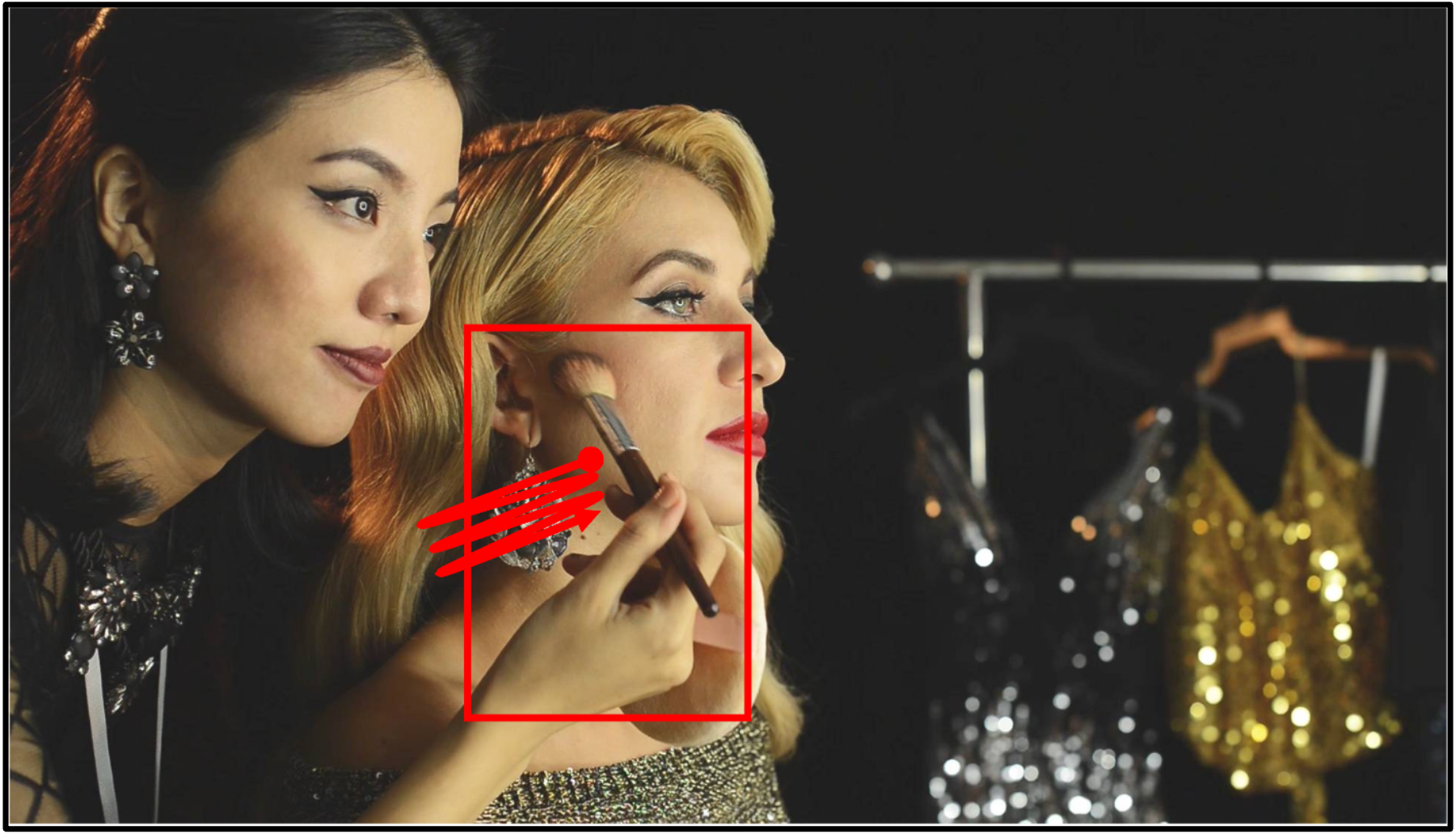}
\end{minipage} &
\begin{minipage}[t]{0.23\textwidth}
  \fcolorbox{black}{white}{\animategraphics[autoplay,loop,width=\linewidth]{24}{images/our_results/3/}{0}{71}}
\end{minipage} &
\begin{minipage}[t]{0.23\textwidth}
  \includegraphics[width=\linewidth]{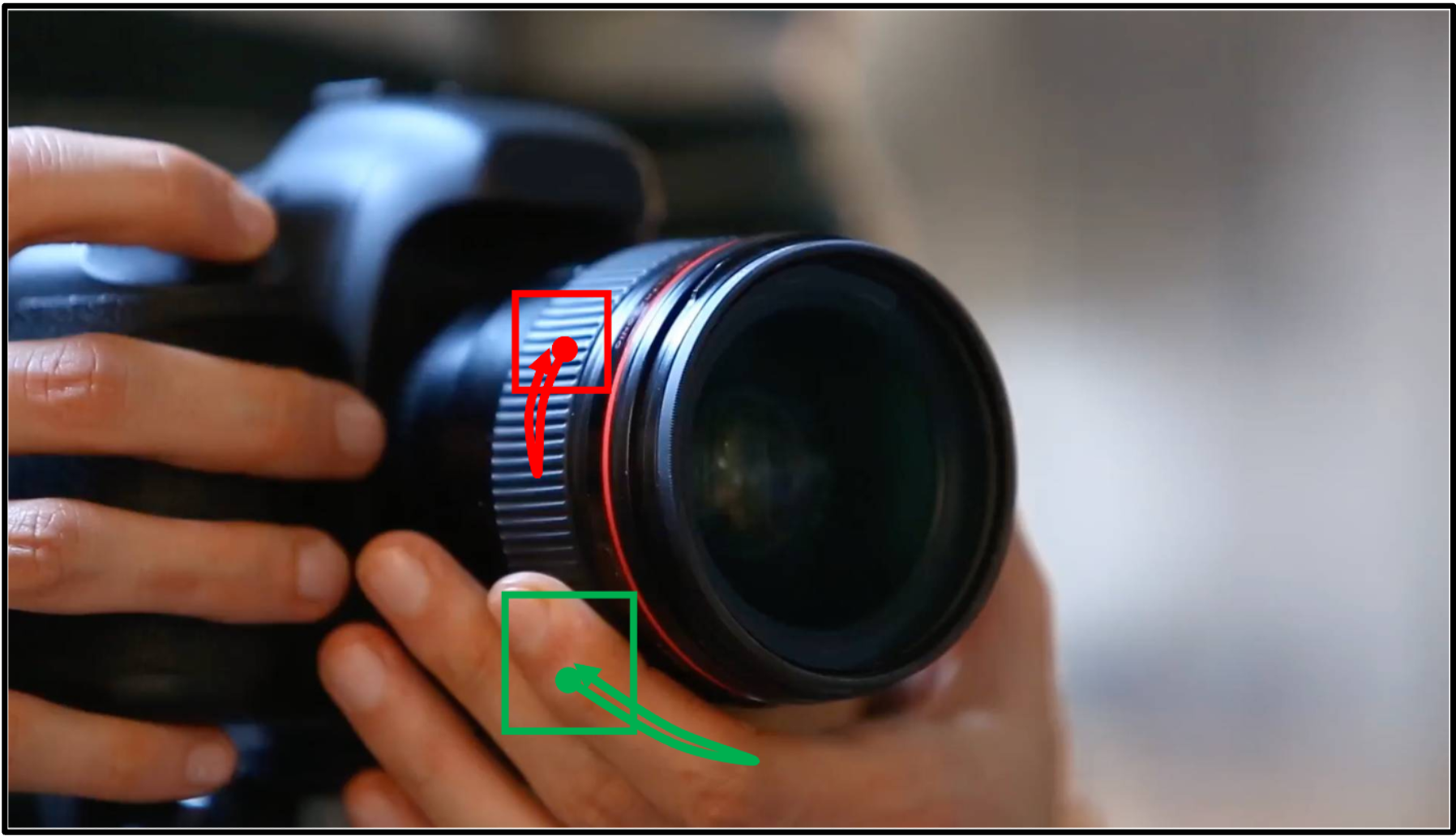}
\end{minipage} &
\begin{minipage}[t]{0.23\textwidth}
  \fcolorbox{black}{white}{\animategraphics[autoplay,loop,width=\linewidth]{24}{images/our_results/4/}{0}{71}}
\end{minipage} \\[-0.3em]
\multicolumn{2}{c}{\small (c) \texttt{`A makeup artist applying makeup brush...'}} &
\multicolumn{2}{c}{\small (d) \texttt{`The person is trying to adjust the...'}} \\
\end{tabular}

\captionof{figure}{\textbf{Our Results.}
Figure shows the robustness of our method to generate very diverse cinemagraphs with different motion patterns, like cat moving its head, or human applying makeup in translation motion. The toy flower petals rotating. (bottom-left) Our method can also simulate complex, realistic hand-object interaction with relatively simple input control signals. \textit{We encourage readers to view the videos in the figure with Acrobat Reader.}}
\label{fig:our_results_rigid}
\end{figure*}

\myparagraph{Single Image Animation.}
Another line of work generates cinemagraphs by starting from a single image and synthesizing periodic motion. One of the earliest efforts is by Chuang \etal who manually specify motion priors for different semantic categories such as water and leaves~\cite{chuang2005animating}. More recent approaches leverage deep networks to predict motion fields~\cite{endo2019animating, logacheva2020deeplandscape, holynski2021animating, mahapatra2022controllable, fan2022simulating, li20233d, bertiche2023blowing, halperin2021endless}. Animating Landscape predicts motion autoregressively and synthesizes videos via backward warping~\cite{endo2019animating}. Holynski \etal instead estimate a single optical-flow field describing motion between consecutive frames and use forward warping to reduce stretching artifacts~\cite{holynski2021animating}. Mahapatra \etal~\cite{mahapatra2022controllable} and Fan \etal~\cite{fan2022simulating} introduce control mechanisms using user-provided masks and directional cues.
A fundamental limitation of these approaches is that they rely heavily on optical-flow prediction, typically in the form of a single flow map~\cite{mahapatra2022controllable, mahapatra2023text, holynski2021animating, halperin2021endless, fan2022simulating}. As a result, they can only animate scenes with repetitive, in-place motions such as water, smoke, or foliage. These methods cannot model the complex non-repetitive dynamics characteristic of rigid objects, including humans, animals, or everyday objects—for example, \textit{“a person applying makeup”} or \textit{“a photographer adjusting a camera lens”} (Figure~\ref{fig:our_results_rigid}).
In contrast, our method is the first to generate cinemagraphs from a single image that works reliably on \textit{general scenes and general motion}, without being limited to fluid elements.

\noindent \textbf{Image-to-Video Generation.}
Recent progress in video foundation models~\cite{blattmann2023stable, xing2023dynamicrafter, chen2023videocrafter1, videoworldsimulators2024, bar2024lumiere, polyak2024movie, yang2024cogvideox, hacohen2024ltx} has enabled highly realistic video generation from a single image and a text prompt. While these models excel at producing diverse and visually compelling videos, their control mechanisms remain largely text-driven. As a result, they cannot provide the fine-grained, spatially localized control over object motion—such as precise trajectories, directional paths, or temporal timing—that cinemagraph creation requires. Moreover, these models are not designed to preserve a static background, nor do they inherently produce seamlessly looping videos, both of which are essential properties of cinemagraphs.
In contrast, our method leverages a pretrained image-to-video diffusion model and finetunes it with first–last frame conditioning and explicit motion signals (bounding-box sequences and sparse point tracks).

\noindent \textbf{Controllable Video Generation.}
Recent progress in controllable video synthesis provides a variety of mechanisms for user-guided motion specification. Existing work typically supports one of three major forms of control: (1) Bounding-box control~\cite{wangboximator, ma2023trailblazer, huang2023factor, li2023trackdiffusion, wu2024motionbooth}, (2) Point-trajectory control~\cite{wu2024draganything, niu2025mofa, wang2024motionctrl, mou2024revideo}, (3) Camera control~\cite{wang2024motionctrl, he2024cameractrl, yu2024viewcrafter, wang2024akira, bahmani2024ac3d}.
More recently, MotionCanvas~\cite{xing2025motioncanvas} enables joint control of both camera and object motion. Instead of relying on explicit camera labels—which are expensive and difficult to obtain—they train the model using sparse point trajectories and interpret such motion during inference as camera movement.
We draw inspiration from these controllable video-generation approaches and examine the problem of cinemagraph generation through the lens of controllable video synthesis, incorporating additional constraints specific to cinemagraphs.
\section{Method}
\label{sec:method}

\newcommand{\vcentered}[1]{\raisebox{0.35\height}{#1}}

\begin{figure*}[h]
\centering
\setlength{\fboxrule}{0.8pt}
\setlength{\tabcolsep}{4pt} 
\setlength{\fboxsep}{0pt}

\begin{tabular}{@{\hspace{0.4em}} c @{\hspace{0.4em}} c @{\hspace{0.4em}} c @{\hspace{0.4em}}}

\begin{minipage}[t]{0.46\textwidth}
  \includegraphics[width=\linewidth]{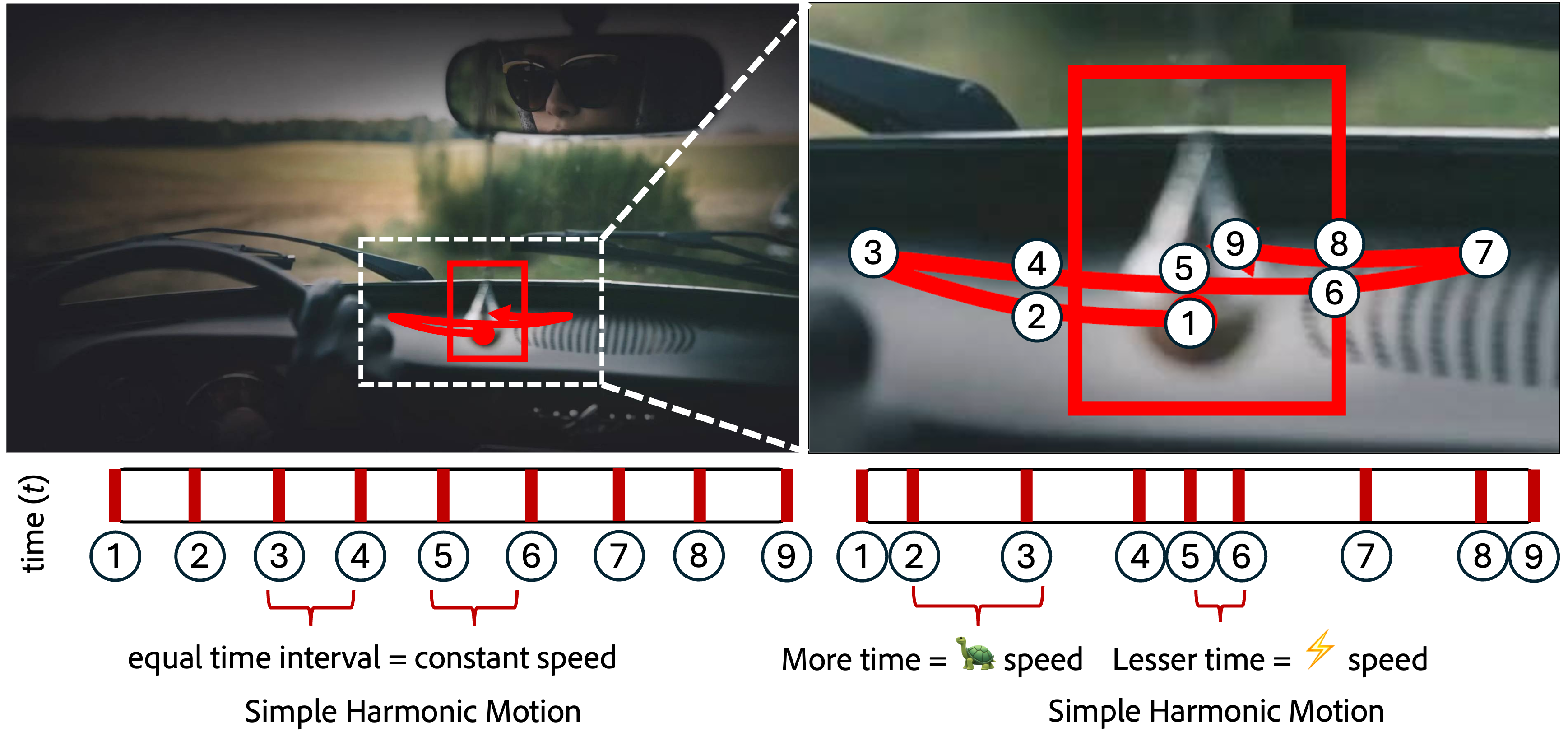}
\end{minipage} &
\begin{minipage}[t]{0.23\textwidth}
  \raisebox{15.5mm}{
  \fcolorbox{black}{white}{\animategraphics[autoplay,loop,width=\linewidth]{24}{images/extra-2/no-timing-control-video/}{0}{71}}
  }
\end{minipage} &
\begin{minipage}[t]{0.23\textwidth}
  \raisebox{15.5mm}{
  \fcolorbox{black}{white}{\animategraphics[autoplay,loop,width=\linewidth]{24}{images/extra-2/timing-control-video/}{0}{71}}
  }
\end{minipage} \\[-4em]
& \multicolumn{1}{c}{\small{\texttt{w/o timing control}}}
& \multicolumn{1}{c}{\small{\texttt{w/ timing control}}} \\
\end{tabular}

\vspace{2.4em}
\rule{0.97\textwidth}{0.6pt}
\vspace{0.4em}

\begin{tabular}{@{\hspace{0.4em}} c @{\hspace{0.2em}} c @{\hspace{1.0em}} c @{\hspace{0.6em}} c @{\hspace{0.4em}}}
\begin{minipage}[t]{0.23\textwidth}
  \includegraphics[width=\linewidth]{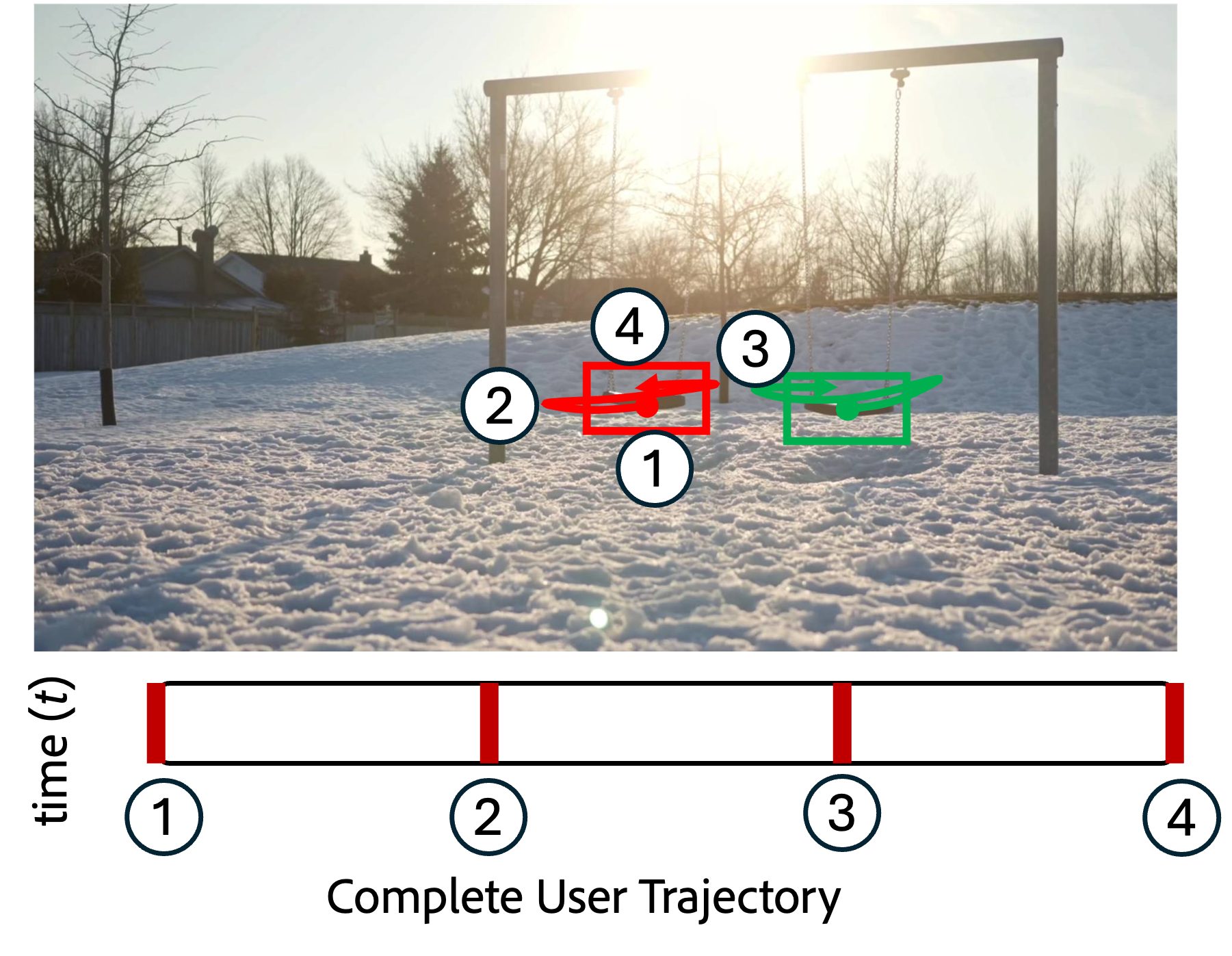}
\end{minipage} &
\begin{minipage}[t]{0.23\textwidth}
  \vcentered{
  \fcolorbox{black}{white}{\animategraphics[autoplay,loop,width=\linewidth]{24}{images/extra-2/full-trajectory-video/}{0}{71}}
  }
\end{minipage} &
\begin{minipage}[t]{0.23\textwidth}
  \raisebox{-2mm}{\includegraphics[width=\linewidth]{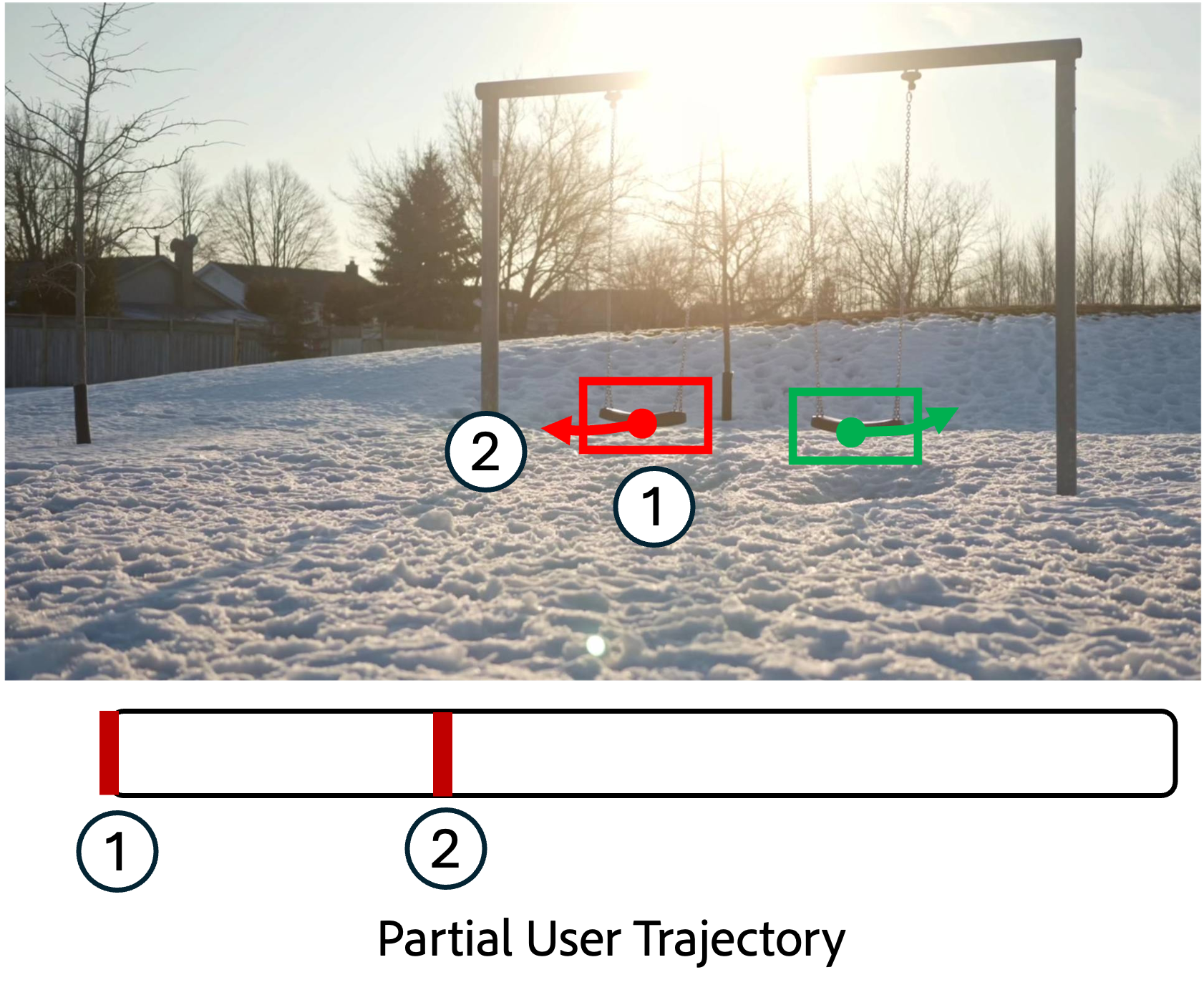}}
\end{minipage} &
\begin{minipage}[t]{0.23\textwidth}
  \vcentered{
  \fcolorbox{black}{white}{\animategraphics[autoplay,loop,width=\linewidth]{24}{images/extra-2/partial-trajectory-video/}{0}{71}}
  } 
\end{minipage} \\[-0.4em]
\end{tabular}

\captionof{figure}{{\bf Fine Grained Controls .} (top) \textbf{Timing controls:} Highlights a scenario where timing control is essential for realistic motion. User intends to simulate the bead osciallting in simple harmonic motion (SHM). Without timing control, i.e., equally space points (top-left) the bead has constant motion, not following SHM, thus looking unrealistic. With timining control, user can assign more time the extrema and lesser time at the minima of SHM, accurately simulating physically realistic motion. (bottom) \textbf{Full vs. Partial motion paths:} Demonstrate 2 scenarios of the same example, where the user can either provide the full motion path (bottom-left) or just the initial motion trajectory (bottom-right) and remaining is automatically generated with our model.}
    \lblfig{extra}
\end{figure*}

Given a user photograph $I\in\mathbb{R}^{H'\times W'\times 3}$, a text prompt $c$, and motion conditions consisting of $N$ bounding boxes $(x_1,y_1,x_2,y_2)_n\in\mathbb{R}^4$ and $M$ point tracks $(x,y)_m\in\mathbb{R}^2$ specified for $T'$ frames, our goal is to synthesize a cinemagraph $V\in\mathbb{R}^{T'\times H'\times W'\times 3}$ of length $T'$.
As discussed in \refsec{intro} and \ref{sec:related_work}, most single‑image-to-cinemagraph methods rely on predicting optical flow and use a single optical‑flow field~\cite{mahapatra2022controllable,mahapatra2023text,holynski2021animating,halperin2021endless,fan2022simulating} to animate the input image. This stationary-flow assumption restricts them to scenes with repetitive, in‑place motions (e.g., water, smoke) and makes them unsuitable for complex, non‑repetitive dynamics of general scene elements such as humans and animals. 
Our approach targets this general setting by leveraging the strong motion priors of large‑scale video diffusion models (VDMs) trained on millions of open-domain videos. We first outline why off‑the‑shelf VDMs are not directly suitable for cinemagraphs in Section \ref{vdm}, then present our adaptations that yield high‑quality, controllable loops from a single input photo. We customize VDMs for cinemagraph generation by (i) enforcing seamless looping (Section \ref{first-last}), and (ii) enabling precise, user‑guided control through motion conditioning on the provided bounding box sequences and point tracks (Section \ref{motion}). Finally, in Section \ref{user-intent} we describe an easy and intuitive way for the user to provide motion control signals to our method.

\subsection{Video Diffusion Models}
\label{vdm}

Video diffusion models (VDMs) are large–scale generative models trained on millions of in‑the‑wild videos. 
Because our input is a single photograph, we focus on image‑to‑video (I2V) models. 
We build upon a pretrained DiT‑based~\cite{peebles2023scalable} I2V system: the input image is encoded by a 3D‑VAE into spatiotemporal latents, the text prompt is encoded by a text encoder, and both are concatenated with noisy tokens and processed by a DiT transformer; a 3D‑VAE decoder then transforms the latent to an RGB video.
We use a variant trained with the flow‑matching objective~\cite{lipman2022flow}. 
Let $X^t$ denote the latent at time $t\!\in\![0,1]$ and $V^t$ are the target velocity along the training path. 
The model learns a velocity field $v_\theta(\cdot,t)$ by minimizing the expected squared error:
\begin{equation}
\mathcal{L}_{\text{FM}}(\theta)
= \min_\theta
\big\|\, V^t - v_\theta(X^t,t) \big\|_2^2 .
\label{eq:fm-loss}
\end{equation}
\noindent We leverage the rich motion priors learned by VDMs trained on massive, diverse video corpora as a starting point toward single‑image‑to‑cinemagraph generation. \\
\noindent\textbf{Cinemagraphs vs. videos.}
Standard I2V VDMs are optimized to produce fully dynamic videos, whereas cinemagraphs contain mostly static scenes with motion confined to selected regions and required to loop seamlessly. 
Bridging this mismatch is non‑trivial: one must preserve a static background, enforce the resulting videos to be looping smoothly, and provide precise, user‑guided motion control for our task of controllable cinemagraph generation. 
The next sections describe how we transform the pretrained VDM to satisfy these cinemagraph‑specific constraints.

\subsection{Time Bridging with Temporal-Boundary Conditioning}
\label{first-last}
\begin{figure*}[h]
\centering
\setlength{\fboxrule}{0.8pt}
\setlength{\fboxsep}{0pt}

\begin{tabular}{@{\hspace{0.0em}} c @{\hspace{0.4em}} c @{\hspace{0.0em}} c @{\hspace{0.0em}} c @{\hspace{0.4em}} c @{\hspace{0.0em}}}
\raisebox{-0.05em}{\includegraphics[width=0.19\textwidth]{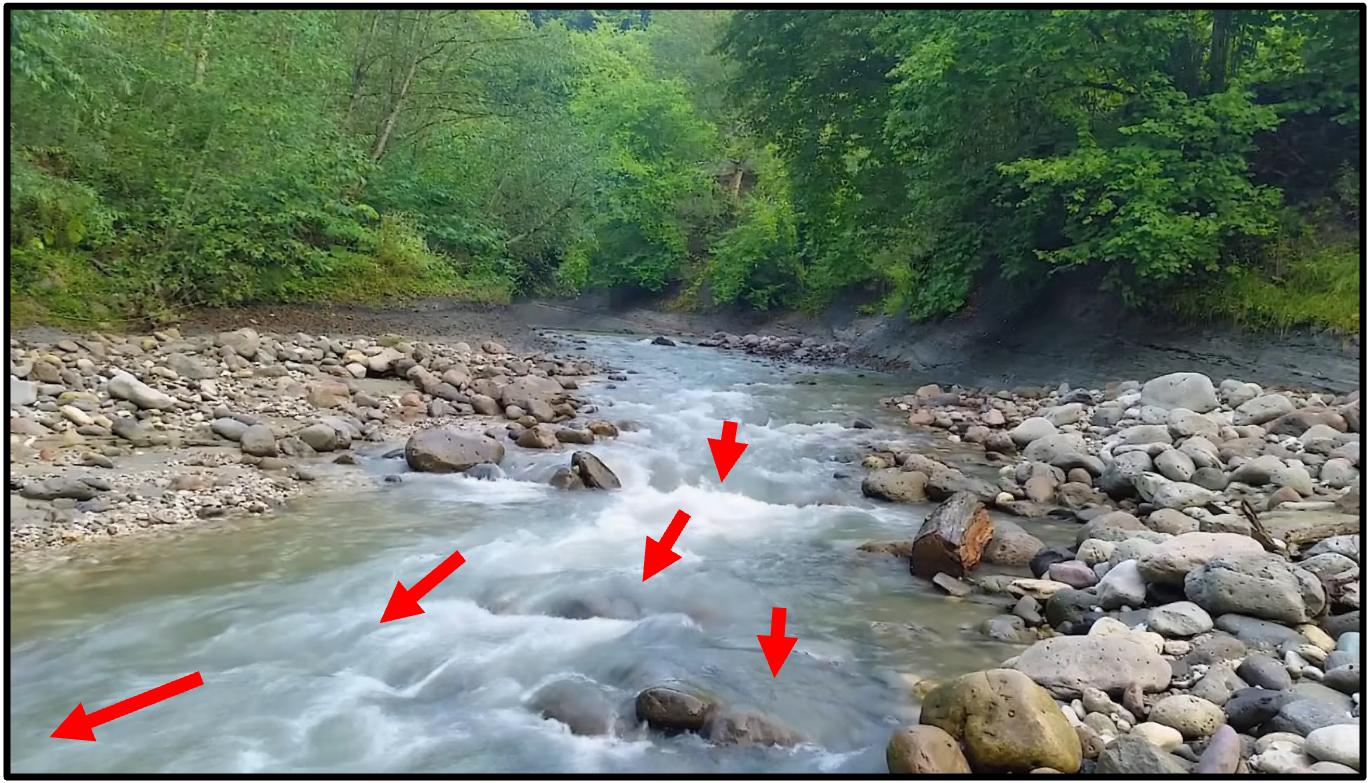}} &
\fcolorbox{black}{white}{\animategraphics[autoplay,loop,width=0.19\textwidth]{24}{images/baseline_comparison/fluid_elements/animating_landscape/}{0}{38}} &
\fcolorbox{black}{white}{\animategraphics[autoplay,loop,width=0.19\textwidth]{24}{images/baseline_comparison/fluid_elements/controllable_animation/}{0}{59}} &
\fcolorbox{black}{white}{\animategraphics[autoplay,loop,width=0.19\textwidth]{24}{images/baseline_comparison/fluid_elements/slr-sfs/}{0}{60}} &
\fcolorbox{black}{white}{\animategraphics[autoplay,loop,width=0.19\textwidth]{24}{images/baseline_comparison/fluid_elements/our_v1p8/}{0}{119}} \\

\small Input Image** &
\small Animating Landscape~\cite{endo2019animating} &
\small Controllable Animation~\cite{mahapatra2022controllable} &
\small SLR-SFS~\cite{fan2022simulating} &
\small Ours
\end{tabular}

\vspace{0.4em}
\rule{0.97\textwidth}{0.6pt}
\vspace{0.4em}

\begin{tabular}{@{\hspace{0.0em}} c @{\hspace{0.4em}} c @{\hspace{0.4em}} c @{\hspace{0.4em}} c @{\hspace{0.4em}} c @{\hspace{0.0em}}}
\raisebox{-0.05em}{\includegraphics[width=0.19\textwidth]{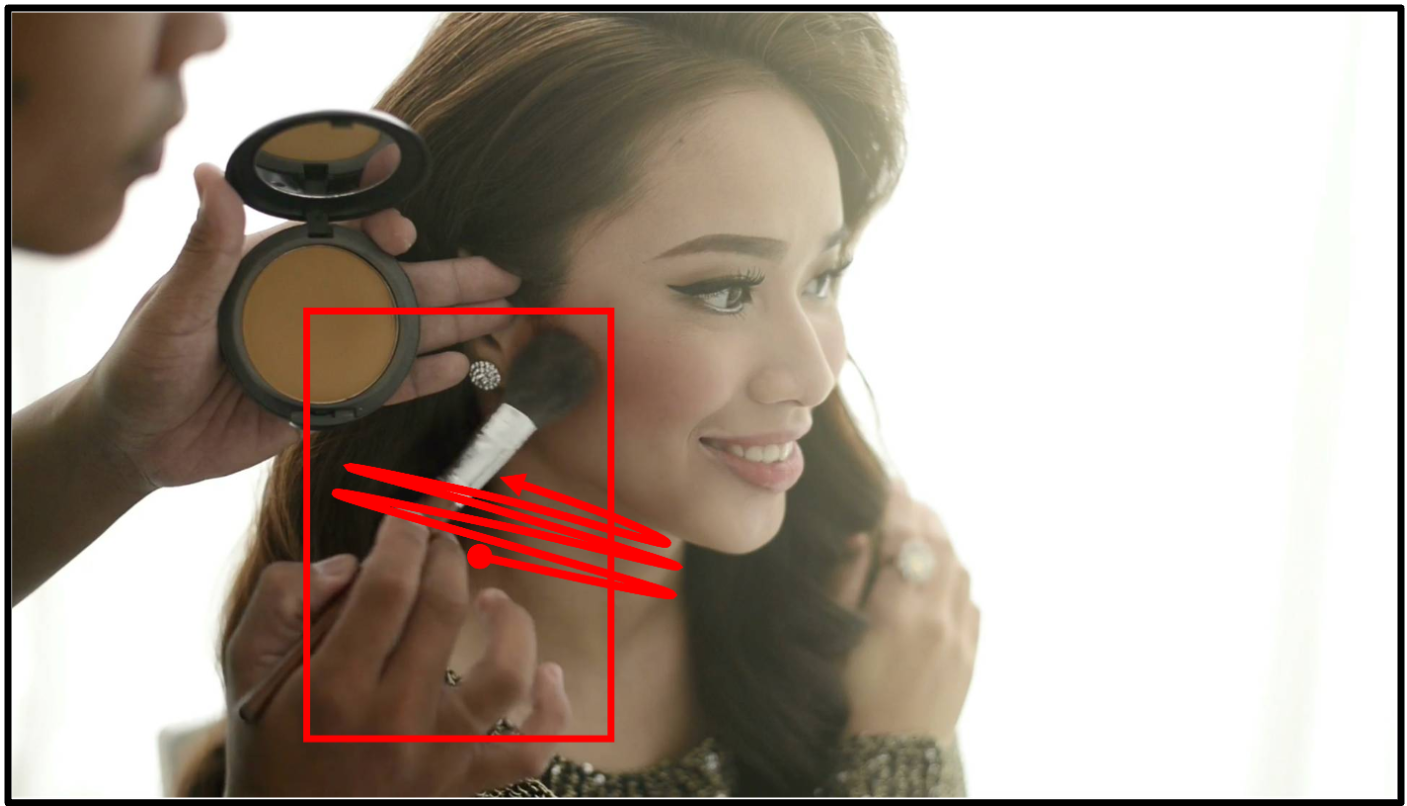}} &
\fcolorbox{black}{white}{\animategraphics[autoplay,loop,width=0.19\textwidth]{8}{images/baseline_comparison/rigid_elements/draganything/}{0}{10}} &
\fcolorbox{black}{white}{\animategraphics[autoplay,loop,width=0.19\textwidth]{24}{images/baseline_comparison/rigid_elements/cogvideox/}{0}{25}} &
\fcolorbox{black}{white}{\animategraphics[autoplay,loop,width=0.19\textwidth]{24}{images/baseline_comparison/rigid_elements/ours_v1p8_nocontrol/}{0}{119}} &
\fcolorbox{black}{white}{\animategraphics[autoplay,loop,width=0.19\textwidth]{24}{images/baseline_comparison/rigid_elements/ours_v1p8/}{0}{71}} \\

\small Input Image** &
\small DragAnything*~\cite{wu2024draganything} &
\small CogVideoX-5B~\cite{yang2024cogvideox} &
\small Ours [only first-last] &
\small Ours
\end{tabular}

\captionof{figure}{\textbf{Comparison with baselines.} Our full method (left) generates cinemagraphs with more realistic motion and accurately follows the input direction signal, both for the case of \textit{'fluid-elements'} cinemagraphs (top) and \textit{'general-domain'} cinemagraphs (bottom). (*) Denotes we apply postprocessing from ~\cite{endo2019animating} to make the generated videos loop. (**) Denotes arrows on inputs for motion representative purpose. \textit{We encourage readers to view the videos in the figure with Acrobat Reader.}}
\label{fig:fluid_elements_comparison}
\end{figure*}

At first glance, cinemagraphs differ from generic videos in that their motion must loop seamlessly and continuously. 
A seemingly straightforward way to adapt VDMs is to train them directly on cinemagraph datasets, but such data are scarce and often low quality. 
Instead, we exploit large‐scale in‑the‑wild video data and train the VDM on the proxy task of time bridging to encourage the model to learn to interpret the temporal connection between two arbitrary point in time.
We achieve this with the temporal-boundary conditioning strategy. 
Concretely, given a training clip $\{I_t\}_{t=0}^{T}$, we encode the boundary frames with the pretrained 3D‑VAE encoder $\mathcal{E}$ from the base DiT to obtain $C_{I_0},C_{I_T}$. 
These tokens are concatenated to the diffusion tokens and provided to the transformer at all timesteps. 
At inference, a cinemagraph becomes a special case by setting the boundary conditions equal to the input photo, i.e., $I_0=I_T=I$, which enforces a seamless loop.
However, temporal-boundary conditioning alone is insufficient for controllable cinemagraphs. 
(i) Text prompts are often ambiguous and can yield incorrect or unexpected motion, especially in cases where users desire to have fine grained motion control; for example, in Figure \ref{fig:fluid_elements_comparison} (bottom), the variant “Ours (temporal-boundary only)” hallucinates an additional person even though the user intent is to move only the hand and brush. 
(ii) In some cases, the model collapses to nearly static frames. 
We present additional failure cases in \website.


\subsection{Motion Control}
\label{motion}
To address the limitations of only using temporal-boundary conditioning for cinemagraph generation and to capture user intent of fine-grained motion control, we further propose to augment the video generative model with object motion conditioning. To choose two forms of conditions for motion control with VDMs:
(1) \textbf{Bounding Boxes: }We capture the object's global motion using bounding boxes, which capture the spatial information of the object. For a video, given coordinates $(x_1, y_1, x_2, y_2)$ for an object denoting the top-left and bottom-right corners, similar to MotionCanvas~\cite{xing2025motioncanvas}, we turn the coordinates into an RGB mask sequence $C_{bbox}^{RGB}\in\mathbb{R}^{T'\times H'\times W'\times 3}$ with each object being color-coded with a unique color. This RGB mask is then encoded by the same 3D-VAE encoder $\mathcal{E}$ to produce a spatiotemporal representation $C_{bbox}=\mathcal{E}(C_{bbox}^{RGB}) $.
(2) \textbf{Sparse Point Tracks: } Point tracks are useful for capturing local object motion. Additionally, a major advantage and our key insight for using point track conditioning for cinemagraphs is that the videos used for training VDMs often contain camera motion.  To prevent undesirable motion in non-target regions during cinemagraph generation, we put static point tracks on those regions to keep them static when generating the cinemagraph. This immensely captures another key aspect of cienmagraphs, where certain regions of the image stay static, in an uncanny but aesthetic manner. Similar to the bounding boxes, we render $N$ point trajectories using color-coded squares into an RGB sequence $C_{traj}^{RGB}\in\mathbb{R}^{T'\times H'\times W'\times 3}$ and use $\mathcal{E}$ to encode them, producing $C_{traj}=\mathcal{E}(C_{traj}^{RGB})$. 
The resulting tokens $C_{motion}=(C_{bbox} \oplus C_{traj})$ are then added to the input noise tokens to the VDM.
The final training objective is given by
\begin{equation}
    \mathcal{L}_{\text{FM}}(\theta)=\min_\theta \| V^t - v_\theta(X^t, t | C_{I_0}, C_{I_T}, C_{motion}, C_{txt}) \|_2^2,
\end{equation}
where $C_{txt}$ is the input prompt tokenized by the text encoder.

\subsection{Capturing User Intent}
\label{user-intent}
Even though during training with motion control, we condition the VDMs for per-frame bounding box and trajectories, during inference, it would be tedious for users to define per-frame motion conditions. Thus, we let the user specify a set of sparse bounding boxes (or trajectories) with timing control, and use spline interpolation to get the in-between dense bounding boxes or trajectories. We provide more details on inference and our demo user interface in \website.

\subsection{Enabling fine-grained controls}
\textbf{Full vs. Partial motion paths}. Our method provides users with full control over the motion pattern they want the selected moving elements to follow by specifying the full, detailed motion paths. However, there are scenarios where users simply want to control coarser motion information, such as the initial direction and speed of the movements. Our system also supports these use cases by enabling motion control with partial motion paths.
For a video sequence of $T$ frames, we allow users to specify the motion path of the target objects/regions for the first $T'\in[1, T]$ frames. If the bounding boxes (or point tracks) are not provided for all $T$ frames, the model is able to infer from the partial input to complete the sequence. This flexibility is enabled by randomly dropping varying amounts of the latter portion of the bounding box (or point tracks) sequences during training. For example, in \reffig{extra} (bottom), we see for the same input example, the user can provide either a full motion sequence (bottom-left) or just a partial sequence (bottom-right), and let the model interpret the remaining motion. Videos corresponding to the inputs are in \website.

\noindent\textbf{Refinement with region controls}. During inference, we fix the point trajectories to be static to disable camera motion (as shown in \reffig{method} (Inference)). While in most cases the static trajectories also prevents motion of the background regions, there are special cases where small, textured regions may still have small local motion. This is likely a consequence of video priors where although many videos contain static content with respect to the \textit{global} scene, there is often small \textit{local} motion in these areas. To prevent motion in non-target regions, users may provide an optional mask which can be used to blend the original image with the generated output.

\noindent\textbf{Timing controls}. The motion paths of the dynamic regions can either be specified on a frame-by-frame basis or by interpolating between a set of anchor bounding boxes placed on the timeline (Section \ref{user-intent}), which is much more easier for user than providing per-frame motion condition. The precise timing and speed of the motions can be controlled by adjusting the anchors on the timeline. One particular example scenario is described in \reffig{extra} (top), where the user wants to generate a cinemagraph of a metallic bead osciallting in a \textit{`simple harmonic motion'}. In this example the user only inputs the bounding boxes at 1...9 discrete intervals, and remaining sequqnce is interpreted with spline interpolation. 
Without timing control the bead would move at an approximately uniform speed.
However, to simulate realistic \textit{`simple harmonic motion'}, user can assign longer dwell times near the extrema and shorter times near the center, thus inducing the characteristic slow–fast–slow \textit{`simple harmonic motion'} profile. Additional examples and videos illustrating the benefits of timing control are provided in \website.

\section{Experiments}
\label{sec:experiments}

\begin{table}[!t]
\centering
\setlength{\tabcolsep}{4pt} 
\scalebox{0.8}{
\begin{tabular}{lcccc} 
\toprule
\textbf{Method} & \textbf{FVD} & \textbf{DT-FVD} & \textbf{KID} & \textbf{FID} \\
\midrule
Animating Landscape~\cite{endo2019animating} & 840.12 & 0.64 & 12.42 & 18.29 \\
Text2Cinemagraph~\cite{mahapatra2023text} & 597.14 & 0.46 & 3.62 & 8.68 \\
\hdashline
Wan2.2-5B*~\cite{wan2025} (pretrained) & 430.03 & 0.63 & 23.56 & 26.27 \\
CogVideoX*~\cite{yang2024cogvideox} & 782.68 & 2.14 & 57.81 & 38.11 \\
\hdashline
Ours (Wan2.2-5B) [only first-last] & \textbf{306.36} & \textbf{0.31} & 4.82 & \textbf{7.14} \\
Ours (internal) [only first-last] & 338.67 & 0.39 & \textbf{3.46} & 7.21 \\
\bottomrule
\end{tabular}
}
\caption{\textbf{Quantitative Evaluation on \textit{'fluid-elements'} (Uncontrollable).} Comparison of our method with baselines using common video generation metrics. Bold numbers indicate the best performance per column. (*) Denotes we apply postprocessing from ~\cite{endo2019animating} to make the generated videos loop.}
\label{tab:natural_uncontrollable}
\end{table}

\myparagraph{Implementation Details.} 
We begin with a pretrained Image-to-Video (I2V) model and first finetune it for first--last frame--conditioned generation. We then extend training to incorporate bounding-box and trajectory control with $\sim8M$ videos. During this stage, we apply two conditioning dropout strategies: (i) randomly dropping between $0$ and $N$ bounding boxes or trajectory points across the sequence, and (ii) dropping all conditioning inputs after the first $K$ frames to enable partial user-specified motion paths (Section~\ref{motion}).
We train two variants of our method: one based on our internal model, finetuned to generate 3-second and 5-second videos at $1080$p resolution, and one based on the Wan2.2-5B~\cite{wan2025} model, finetuned for 3-second videos at $544$p resolution. All experiments are conducted on four A100 nodes. Additional architectural and training details are provided in \website.
\\
\myparagraph{Evaluation.} 
Unlike prior works~\cite{endo2019animating, holynski2021animating, mahapatra2022controllable, mahapatra2023text, fan2022simulating}, which rely on optical-flow-based warping and are therefore limited to natural scenes with repeating textures (e.g., water, smoke), our method imposes no such restrictions. Nonetheless, because our approach is also applicable to these settings, we conduct two types of evaluations. 
(i) We compare against prior methods on cinemagraph generation for natural scenes such as rivers, lakes, and coastal water dynamics; we refer to this setting as \textit{`Fluid-Elements Cinemagraphs'}.  
(ii) We additionally evaluate on a significantly more challenging regime involving arbitrary dynamic content, including humans, objects, and animals; we refer to this setting as \textit{`General-Domain Cinemagraphs'}. 

\begin{table}[!t]
\centering
\setlength{\tabcolsep}{4pt} 
\scalebox{0.8}{
\begin{tabular}{lcccc} 
\toprule
\textbf{Method} & \textbf{FVD} & \textbf{DT-FVD} & \textbf{KID} & \textbf{FID} \\
\midrule
\multicolumn{5}{l}{\textbf{[1 Hint Point]}} \\ 
\midrule
Animating Landscape~\cite{endo2019animating} & 795.02 & 0.58 & 11.31 & 17.44 \\
SLR-SFS~\cite{simo2016learning} & 402.34 & 0.33 & 6.87 & 9.28 \\
Controllable Animation~\cite{mahapatra2022controllable} & 434.61 & 0.38 & 7.22 & 10.21 \\
\hdashline
Ours (Wan2.2-5B) & \textbf{269.38} & \textbf{0.29} & 4.62 & \textbf{7.08} \\
Ours (internal) & 286.34 & 0.3 & \textbf{3.11} & 7.24 \\
\midrule 
\multicolumn{5}{l}{\textbf{[5 Hint Points]}} \\ 
\midrule
Animating Landscape~\cite{endo2019animating}  & 773.71 & 0.56 & 11.97 & 17.61 \\
SLR-SFS~\cite{simo2016learning} & 379.47 & 0.35 & 7.16 & 9.24 \\
Controllable Animation~\cite{mahapatra2022controllable} & 414.78 & 0.38 & 7.92 & 10.06 \\
\hdashline
Ours (Wan2.2-5B) & \textbf{256.73} & \textbf{0.29} & 4.62 & \textbf{7.09} \\
Ours (internal) & 272.38 & \textbf{0.29} & \textbf{3.33} & 7.17 \\
\bottomrule
\end{tabular}
}
\caption{\textbf{Quantitative Evaluation on \textit{'fluid-elements'} (Controllable).} Comparison of our method with baselines using common video generation metrics under single and multi-trajectory-point control settings. Bold numbers indicate the best performance per column. (*) Denotes we apply postprocessing from ~\cite{endo2019animating} to make the generated videos loop.}
\label{tab:natural_controllable_traj}
\end{table}

\begin{table*}[!t]
\centering
\setlength{\tabcolsep}{4pt}
\scalebox{0.72}{
\begin{tabular}{lcccccc}
\toprule
Method & Aesthetic Quality & Imaging Quality & Temporal Flickering & Motion Smoothness & Background Consistency & Subject Consistency \\
\midrule
CogVideoX*~\cite{yang2024cogvideox} & 0.5353 & 0.6419 & 0.9888 & 0.9904 & 0.9741 & 0.9787 \\
Wan2.2-5B*~\cite{wan2025} (pretrained) & 0.5442 & 0.6018 & 0.9881 & 0.9922 & 0.9598 & 0.9605 \\
\hdashline
\rule{0pt}{10pt}DragAnything*~\cite{wu2024draganything} & 0.4986 & 0.5535 & 0.9674 & 0.9789 & 0.9613 & 0.9474 \rule[-5pt]{0pt}{0pt} \\ 
\hdashline
Ours (Wan2.2-5B) [only first-last] & 0.5897 & 0.6756 & 0.9950 & 0.9959 & 0.9746 & 0.9941 \\
Ours (internal) [only first-last] & 0.5872 & 0.6615 & 0.9899 & 0.9929 & 0.9685 & 0.9566 \\
Ours (Wan2.2-5B) & \textbf{0.5998} & 0.6762 & 0.9955 & \textbf{0.9964} & 0.9800 & \textbf{0.9868} \\
Ours (internal) & 0.5997 & \textbf{0.6771} & \textbf{0.9965} & 0.9959 & \textbf{0.9849} & 0.9858 \\
\bottomrule
\end{tabular}
}
\caption{\textbf{Quantitative evaluation on \textit{`Rigid-Body Cinemagraphs'}.} 
Comparison of our methods against baseline Image-to-Video and controllable generation models across multiple VBench~\cite{huang2024vbench} metrics. Bold numbers indicate the best performance per column. (*) Denotes we apply postprocessing from ~\cite{endo2019animating} to make the generated videos loop.}
\label{tab:custom_controllable}
\end{table*}

\subsection{Fluid-Elements Cinemagraphs}
\myparagraph{Evaluation Dataset and Metrics.} Following prior works~\cite{endo2019animating, holynski2021animating, mahapatra2022controllable, mahapatra2023text, fan2022simulating}, we evaluate our method and baselines on 162 nature-scene videos with fluid motion provided by ~\cite{holynski2021animating}, predominantly containing rivers, lakes, ocean etc. We evaluate them under two different settings: i) uncontrollable generation - only image as inputs, ii) controllable generation - with input image and direction guidance. Following~\cite{li2024_GenerativeImageDynamics} we evaluate the results on FVD~\cite{ge2024content, unterthiner2018towards}, DTFVD~\cite{dorkenwald2021stochastic}, FID~\cite{kynkaanniemi2019improved,parmar2022aliased, heusel2017gans} and KID~\cite{binkowski2018demystifying}. The text-prompts corresponding to these examples are generated with GPT-5, and masks (of region to move in generated cinemagraphs) are generated with ODISE~\cite{xu2023open} following Text2Cinemagraph~\cite{mahapatra2023text}.

\noindent\textbf{Direction Guidance.} Under the controllable generation setting, we evaluate our method and baselines with 1 and 5 direction inputs (hints). to get these hints automatically, we following Controllable Animation~\cite{mahapatra2022controllable}, and perform K-Means clustering on the ground-turth optical flow in the dataset and identify 5 (or 1) clusters. We then propagate the optical flow value of these hints in the masked region of the image. These are treated as guidance optical flow maps for baselines. For our method we use this guidance optical flow map to generate per-frame point tracks. For evaluation purpose, we use point tracks on $10x10$ grid, similar to our training setting, thought our method is robust to different number of point tracks (examples in \website).

\myparagraph{Baselines.} 
We evaluate our method under two broad categories.  
(\(\mathbf{I}\))~\textbf{Uncontrollable Generation.} 
We compare against: 
(i) \textit{Optical flow-based methods} --- Animating Landscape~\cite{endo2019animating} and Text2Cinemagraph~\cite{mahapatra2023text}; and 
(ii) \textit{Image-to-Video models} --- CogVideoX-5B~\cite{yang2024cogvideox} and Wan2.2-5B~\cite{wan2025} (pretrained).  
We compare them against our model variant trained only with first–last frame conditioning for comparison.  
(\(\mathbf{II}\))~\textbf{Controllable Generation.} 
We compare with controllable optical flow-based approaches, including Animating Landscape (control variant)~\cite{endo2019animating}, SLR-SFS~\cite{simo2016learning}, and Controllable Animation~\cite{mahapatra2022controllable}. Following Controllable Animation~\cite{mahapatra2022controllable}, we evaluate these models using both one and five directional hints.  
Since Image-to-Video models do not inherently generate looping videos, we apply a post-processing procedure similar to that of Endo \textit{et al.}~\cite{endo2019animating} to produce seamless loops.

\myparagraph{Effectiveness on Uncontrollable Generation} From Table~\ref{tab:natural_uncontrollable} shows that our method—using both Wan2.2‑5B and our internal model—outperforms the baselines on all metrics --- FVD, DT‑FVD, KID, and FID. This improvement is largely due to leveraging VDMs, which yield more realistic motion than optical‑flow–based warping baselines. Because Wan2.2‑5B and CogVideoX produce non‑looping videos, we apply post‑processing to enforce looping; this can degrade frame quality, introducing blending artifacts, especially when the generated videos contain camera motion. We provide video results in \website showing side‑by‑side comparisons, where our results exhibit noticeably more realistic motion. Additonally, our method with Wan2.2-5B backbone performs better than internal model which can be attributed to different learned priors.

\myparagraph{Effectiveness on Controllable Generation.} Similar to the scenario of Uncontrollable Generation, from Table~\ref{tab:natural_controllable_traj}, we find that for both 1-hint and 5-hints, our methods perform better than baselines in terms of all metrics. We see from Figure~\ref{fig:fluid_elements_comparison} (top) that optical flow-based baselines can generate unrealistic motion (Animating Landscape~\cite{endo2019animating}), or have holes in the generated frames due to incorrect optical flow estimation or inability of the model to fill holes introduced by optical-flow warping (SLR-SFS~\cite{fan2022simulating}). In contrast, our method produces realistic water motion.

\subsection{General-Domain Cinemagraphs}
\myparagraph{Evaluation Dataset and Metrics.} 
For evaluating \textit{`General-Domain Cinemagraphs'}, we manually curate a dataset consisting of 50 images, text prompts, along with corresponding bounding-box and trajectory annotations. The dataset spans a diverse range of scenes, including humans, animals, and inanimate objects. 
We evaluate our method and the baselines using multiple dimensions of VBench~\cite{huang2024vbench}, specifically: \textit{Aesthetic Quality}, \textit{Imaging Quality}, \textit{Temporal Flickering}, \textit{Motion Smoothness}, \textit{Background Consistency}, and \textit{Subject Consistency}. For each sample, we conduct evaluations with three random seeds to account for the generation variability.

\myparagraph{Baselines.} 
We compare our method against three categories of baselines:  
(i) Image-to-Video models: CogVideoX-5B~\cite{yang2024cogvideox} and Wan2.2-5B~\cite{wan2025} (pretrained);  
(ii) Trajectory-controlled Image-to-Video models: DragAnything~\cite{wu2024draganything}; and  
(iii) Our model variants trained with only first–last frame conditioning.  
For categories (i) and (iii), we use detailed text prompts derived from the corresponding motion conditions. Since models in categories (i) and (ii) do not inherently generate looping videos, we apply a post-processing procedure similar to that of Endo \textit{et al.}~\cite{endo2019animating} to produce seamless loops.

\begin{figure}[t!]
    \centering
    \includegraphics[width=\linewidth]{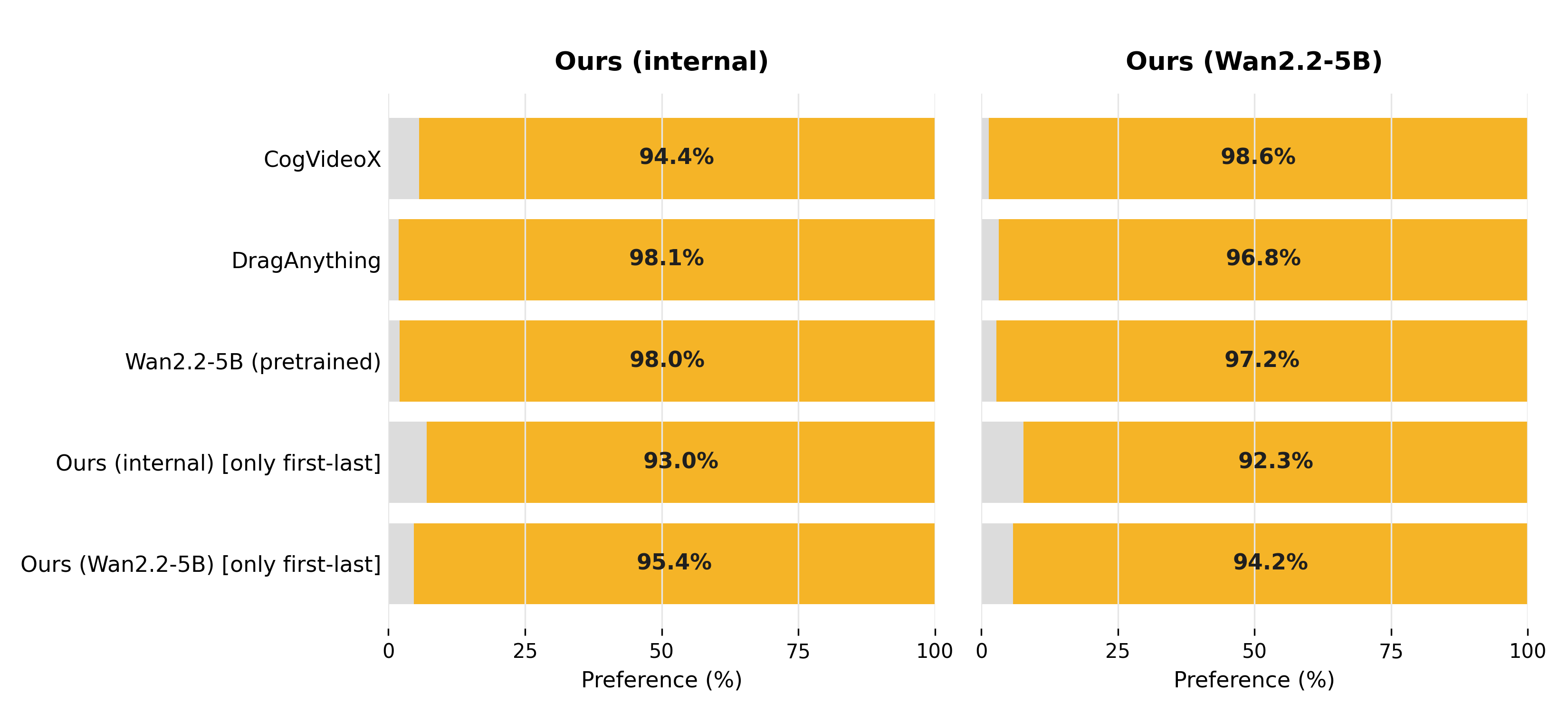}
    \caption{\camera{{\bf User Study.} Figure shows user study of our method, with both internal model and Wan2.2-5B backbone. For both cases, users heavily favour cinemagraphs gernerated by our method compared to the baselines. Bright-yellow denotes percentage our methods is preferred over baselines.}}
    \lblfig{user-study-bar}
\end{figure}

\myparagraph{User-Study.} We also conduct a user study to assess the quality of generated cinemagraphs. We ask the Amazon MTurk participants to choose which animation they prefer, based on following the bounding box and trajectory control. We perform paired tests, where the users are asked to compare two videos, one generated by our method and the other by one of the baselines. The study was conducted with 50 annotators, where each paired comparison was annotated by three annotators. More details about the User Study are in \website.

\myparagraph{Effectiveness on \textit{`General-Domain Cinemagraphs'}.} From Table~\ref{tab:custom_controllable}, we see that our method, with both Wan2.2-5B and internal model, outperforms all baselines and our method with only first- and last-frame conditions for controllable cinemagraph generation. Specifically, from Figure~\ref{fig:fluid_elements_comparison} (bottom), DragAnything and CogVideoX have blending artifacts due to post-processing to make them loop. Our method, with only first- and last-frame conditions, unexpectedly introduces a human, even though it was not expressed in the input user intent. Our method generates the cinemagraph, precisely capturing the user input. Figure~\ref{fig:user-study-bar} shows that users extensively prefer cinemagraphs generated by our method over baselines by a huge margin. Our method can generate cinemagraphs for diverse scenes, like animals (cat), humans, objects, with different motion patterns like translation (cat head and human applying makeup), rotation (flower), and can capture object interaction accurately (Figure~\ref{fig:user-study-bar} (bottom-right).

\section{Conclusion and Limitations}
\label{sec:conclusion}

This paper presented \dreamloop, a controllable cinemagraph generation technology. It empowers a user to animate any image content with intuitive control of object motion when creating cinemagraphs from a single photo. \dreamloop is built upon a pre-trained image-to-video model and fine-tuned with widely available regular videos without cinemagraphs that are difficult to obtain at scale. The two key enabling innovations of our method are temporal bridging that enables the model to generate loop animation and motion conditioning that allows a user to animate desirable content while keeping the rest of image content static. As demonstrated in the experiments, this technology significantly simplifies high-quality cinemagraph production. \\
\noindent\textbf{Limitations.} 
Even though our method can generate high-quality cinemagraphs capturing user motion guidance with precise timing control, there are still some failure cases:
(1) Because our method is built on top of VDMs, occasionally generated cinemagraphs can have morphing artifacts or implausible motion common to VDMs. (2) Even though we provide static point tracks for regions with no user-guided motion, certain generations can still have in-place small motion. Though they can be corrected with blending using masks. (3) In scenarios when the user-provided direction annotations may be conflicting with each other or physically implausible, the model might ignore one or more motion conditions in the generated cinemagraphs, which may be unsatisfactory for the user. We provide more such cases and video examples in \website.


{
    \small
    \bibliographystyle{ieeenat_fullname}
    \bibliography{main}
}


\clearpage
\appendix

\section{Training Details.}
\noindent\textbf{Training Dataset.} Our training dataset consists of $\sim 8$ Million videos annotated with bounding boxes and sparse point tracks, on a $10x10$ grid. The bounding boxes were extracted from ground-truth video using DEVA~\cite{cheng2023tracking} and Qi~et~al.~\cite{Qi_2023_EntitySeg}. Point tracks were computed using optical flow. We first compute both forward and backward optical flows for the video using RAFT~\cite{teed2020raft}, and only keep those point trajectories for which cyclic consistency holds with forward and backward flows. Sparse point tracks are sampled on a uniform $10x10$ grid.

\noindent\textbf{Control Condition Details.} We condition the model with both bounding boxes and trajectories:
(1) Bounding Box: From all the bounding box sequences for a sample video, we select up to three bounding boxes per iteration. The selection is random with \texttt{multinomial} sampling based on the area of the bounding box (with larger having more probability) for 80\% and completely random for 20\%. The selected boxes are rendered to frames of the same shape as the training video, with a black background. We use one of the RGB channels to encode the bounding boxes; hence, we train with a maximum of 3 bounding boxes. Note: that we can train with more bounding boxes with a different color scheme. This conditioning mechanism is not the primary focus of the work. 
(2) Sparse Point Tracks: Similar to bounding boxes, we render the point tracks on frames of the same shape as the training video with a black background as a square of 10 pixels. The color for each point track is based on its $(x,y)$ coordinate. During training, we randomly drop 1...N point tracks. 
\noindent Both bounding boxes and point trajectories, once rendered to frames, are passed through the pretrained 3DVAE and added to the noisy latent after patchification.
For both conditioning mechanisms, during training, with 50\% probability, we randomly both bounding box and point track conditioning after K frames, where K is sampled uniformly between [0,N], where N is the total number of frames. We do this to simulate the scenario where the user inputs a partial motion path. 

\noindent\textbf{Training Procedure.} We finetune our models on 4 A100 nodes, consisting of 8 GPUs each, starting from a base Image-to-video model. For Wan2.2-5B~\cite{wan2025}, we finetune the model at 544p resolution videos of length 3s (73 frames), starting from the TI2V model. We first finetune the model with first-last-frame conditioning for 10K iterations. The last frame is concatenated in the token dimension as clean latent (similar to what Wan2.2-5B does for Image-to-Video training with the first frame). We further add bounding box and point track conditioning and fine-tune the model for an additional 10K iterations. We use the same learning rate, optimizer, and noise scheduler used in the original model training. For our method's variant with our internal model, we follow a similar procedure, except that the base model is finetuned for 1080p videos of duration 3s and 5s (73 and 121 frames, respectively).

\begin{figure}[t!]
    \centering
    \includegraphics[width=\linewidth]{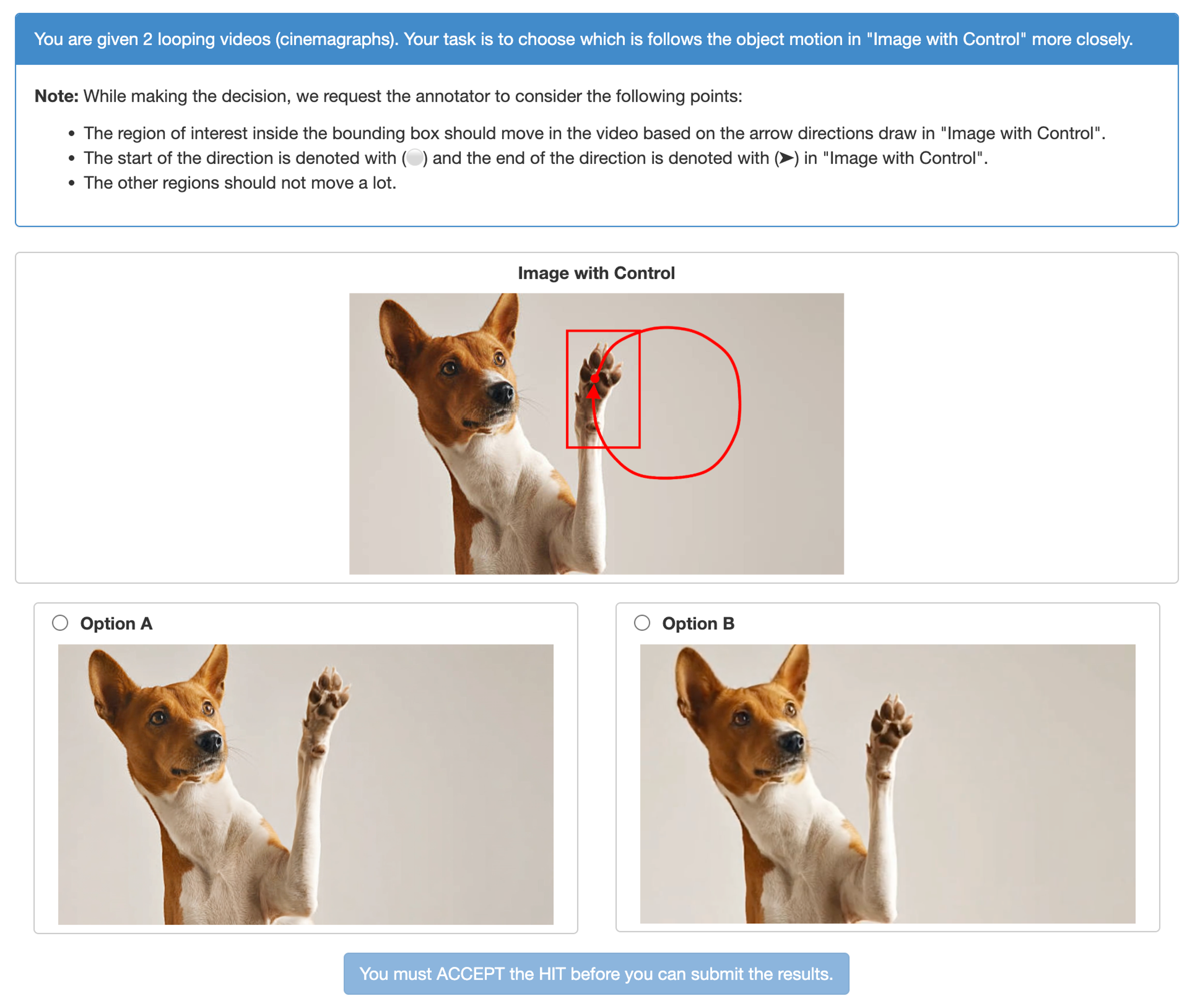}
        \caption{{\bf User Study.} The figure shows the user study setup of our method on Amazon Mechanical Turk. Annotators are shown these instructions and the input image with reference motion path and asked which looping video they find to be better, following the representative motion path.}
    \lblfig{user-study-supp}
\end{figure}

\section{User Study.}
We conduct a user study on Amazon Mechanical Turk (AMT) using a set of 50 controllable cinemagraph videos. The study involves 50 unique annotators, and each video pair is evaluated by three independent workers. \reffig{user-study-supp} illustrates the AMT interface used for the pairwise comparisons.
Given an input image and its corresponding motion-control specification overlaid on the image, annotators are asked to select `which of the two generated videos better follows the prescribed motion control'. To avoid confusion arising from unfamiliar terminology, we replace the word cinemagraph with the more widely understood term looping video in all user-facing instructions.
We conduct two separate user studies:
(1) comparing our internal model against all baselines, and
(2) comparing Wan2.2-5B against the same set of baselines.

\end{document}